\RequirePackage{fix-cm}

\documentclass[twocolumn]{svjour3}

\smartqed  

\usepackage{graphicx}
\usepackage{amsmath,amssymb}

\usepackage{url}
\usepackage{bm}
\usepackage{booktabs}
\usepackage{multirow}
\usepackage{widetext}
\usepackage{microtype}

\newcommand{\Argmax}{\mathop{\rm argmax}}

\newcommand{\mb}[1]{\mathbf{#1}}

\newcommand{\zeros}{\mathbf{0}}

\newcommand{\eg}{\textit{e.g.}}

\newcommand{\ie}{\textit{i.e.}}
\newcommand{\etal}{\textit{et~al.\@}}

\newcommand{\eref}[1]{Eq.~(\ref{#1})}
\newcommand{\tref}[1]{Table~\ref{#1}}
\newcommand{\fref}[1]{Fig.~\ref{#1}}
\newcommand{\sref}[1]{Section~\ref{#1}}

\newcommand{\notapplicable}{-}

\begin{document}

\allowdisplaybreaks[4]

\title{
MirrorNet: A Deep Bayesian Approach to Reflective 2D Pose Estimation from Human Images
}

\author{
    Takayuki~Nakatsuka${}^1$ \and
    Kazuyoshi~Yoshii${}^2$ \and
    Yuki~Koyama${}^3$ \and
    Satoru~Fukayama${}^3$ \and
    Masataka~Goto${}^3$ \and
    Shigeo~Morishima${}^4$
}

\authorrunning{Nakatsuka et al.} 

\institute{
$^1$Takayuki Nakatsuka \at
    Waseda University,
    Shinjuku-ku, Tokyo, Japan. \\
    TEL/FAX: +81-3-5286-3510 \\
    \email{t59nakatsuka@fuji.waseda.jp}
\and
$^2$Kazuyoshi~Yoshii \at
    Kyoto University,
    Sakyo-ku, Kyoto, Japan.
\and
$^3$Yuki Koyama, Satoru Fukayama and Masataka Goto \at
    National Institute of Advanced Industrial Science and Technology (AIST),
    Tsukuba, Ibaraki, Japan.
\and
$^4$Shigeo Morishima \at
    Waseda Research Institute for Science and Engineering,
    Shinjuku-ku, Tokyo, Japan.
}

\date{Received: date / Accepted: date}

\maketitle
\begin{abstract}
This paper proposes a statistical approach
 to 2D pose estimation from human images.
The main problems with the standard supervised approach, which is
 based on a deep recognition (image-to-pose) model,
 are that it often yields anatomically implausible poses,
 and its performance is limited by the amount of paired data.
To solve these problems,
 we propose a semi-supervised method
 that can make effective use of images with and without pose annotations.
Specifically,
 we formulate a hierarchical generative model of poses and images
 by integrating a deep generative model of poses from pose features with
 that of images from poses and image features.
We then introduce a deep recognition model that infers poses from images.
Given images as observed data,
 these models can be trained jointly in a hierarchical variational autoencoding
 (image-to-pose-to-feature-to-pose-to-image) manner.
The results of experiments show
 that the proposed reflective architecture makes
 estimated poses anatomically plausible,
 and the performance of pose estimation improved
 by integrating the recognition and generative models
 and also by feeding non-annotated images.
\begin{figure}[!tb]
\centering
\includegraphics[width=.98\linewidth]{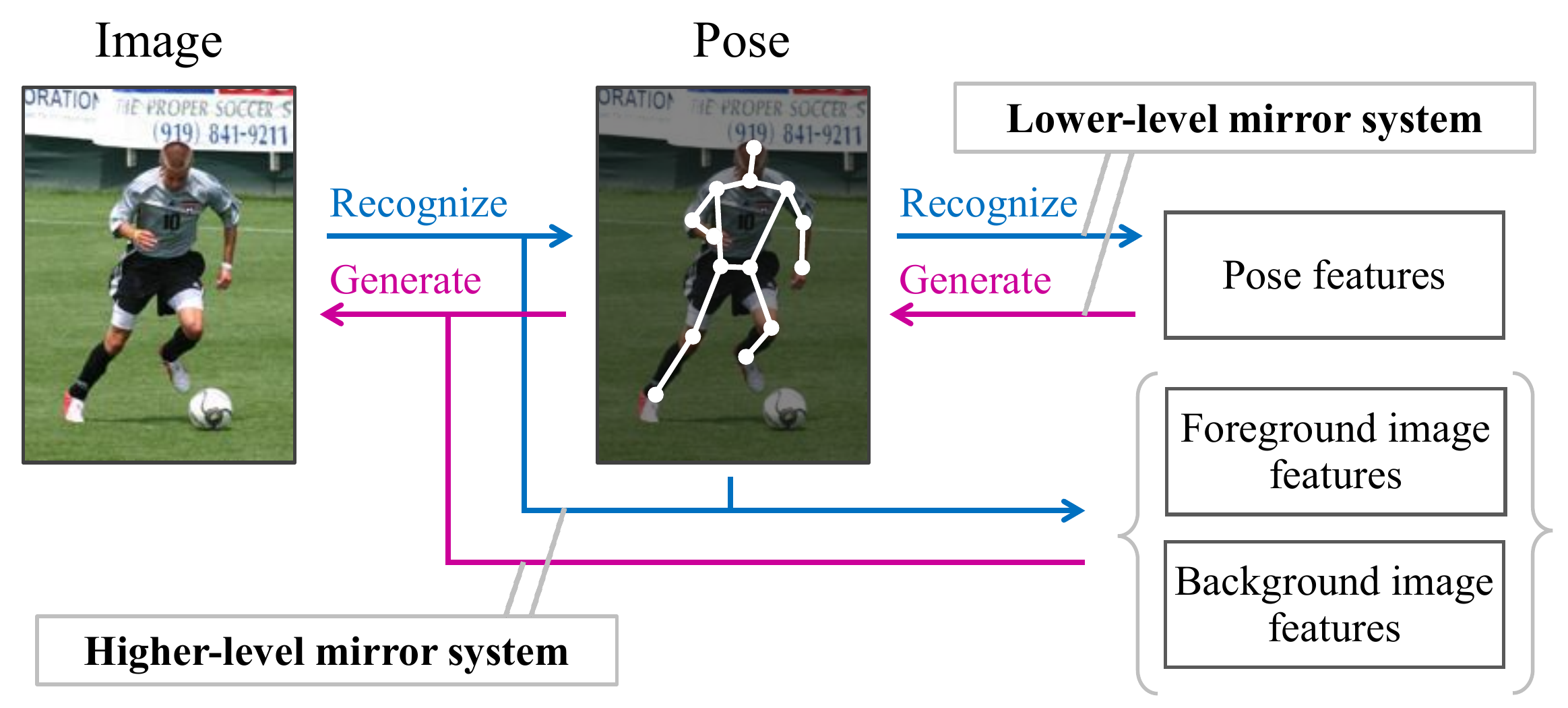}
\caption{
An overview of MirrorNet, which consists of
 generative models of poses and images from latent features
 and recognition models of poses and latent features from images.
The latent features consist of primitives (pose features),
 appearances (foreground image features),
 and scenes (background image features).
A higher-level image-to-pose-to-image mirror system is integrated
 with a lower-level pose-to-feature-to-pose mirror system in a hierarchical Bayesian manner, which enables the unsupervised learning of images without pose annotations
}
\label{fig:concept}
\end{figure}
\end{abstract}


\section{Introduction}
\label{sec:introduction}
Human beings understand the essence of things by abstraction and embodiment.
As Richard P. Feynman, the famous physicist, 
 stated, ``What I cannot create, I do not understand''~\cite{hawking2001universe},
 abstraction and embodiment are two sides of the same coin.
Our hypothesis is
 that such a bidirectional framework plays a key role
 in the brain 
 process of recognizing human poses from 2D images,
 inspired by the \textit{mirror neuron system} or \textit{motor theory}
 known in the field of cognitive neuroscience~\cite{iacobani2007mirror}.
In this paper,
 we focus on the estimation of the 2D pose (joint coordinates) of a person in an image,
 inspired by the human mirror system.

The standard approach to 2D pose estimation
 is to train a deep neural network (DNN)
 that maps an image to a pose in a supervised manner
 by using a collection of images with pose annotations~\cite{toshev2014deeppose,tompson2014joint,newell2016stacked,wei2016convolutional,belagiannis2017recurrent,yang2017learning,sun2019deep}.
Toshev and Szegedy~\cite{toshev2014deeppose}
 pioneered a method called DeepPose
 that uses a DNN consisting of convolutional and fully connected layers
 for the nonlinear regression of 2D joint coordinates from images.
Instead of directly using 2D joint coordinates as target data,
 Thompson~\etal~\cite{tompson2014joint}
 proposed a heatmap representation
 that indicates the posterior distribution of each joint over pixels.
This representation has commonly been used
 in many state-of-the-art methods of 2D pose estimation
 \cite{newell2016stacked,wei2016convolutional,belagiannis2017recurrent,yang2017learning,sun2019deep}.
Note that all these methods focus only on the recognition part of the human mirror system.

Such a supervised approach based on \textit{image-to-pose} mapping
 has two major drawbacks.
First,
 the anatomical plausibility of estimated poses is not taken into account.
To mitigate this problem,
 the positional relationships between adjacent joints have often been considered
 \cite{lifshitz2016human,bulat2016human,chu2017multi,tang2018deeply,nie2018human},
 and error correction networks
 \cite{carreira2016human,chen2018cascaded}
 and adversarial networks
 \cite{chen2017adversarial,chou2018self}
 have been used in a heuristic manner.
Second,
 the performance of the supervised approach
 is limited by the amount of paired pose-image data.
To overcome this limitation,
 data augmentation techniques~\cite{peng2018jointly}
 and additional metadata~\cite{ukita2018semi} have been utilized.
A unified solution to these complementary problems, however, remains an open question.

In this paper,
 we propose a hierarchical variational autoencoder (VAE) called \textit{MirrorNet}
 that consists of higher- and lower-level mirror systems (\fref{fig:concept}).
Specifically,
 we formulate a probabilistic latent variable model
 that integrates a deep generative model of poses from pose features
 (called \textit{primitives})
 with that of images from poses and foreground and background features
 (called \textit{appearances} and \textit{scenes}).
To estimate poses, pose features, and image features from given images
 in the framework of amortized variational inference (AVI)~\cite{kingma2014autoencoding},
 we introduce a deep recognition model of pose features from poses,
 that of foreground and background image features from poses and images,
 and that of poses from images.
These generative and recognition models can be trained jointly
 even from non-annotated images.

A key feature of our semi-supervised method
 is to consider the anatomical \textit{fidelity} and \textit{plausibility}
 of poses in the estimation process.
To make use of both annotated and non-annotated images,
 our method constructs an \textit{image-to-pose-to-image} reflective model
 (\ie, a higher-level mirror system for image understanding)
 by connecting the \textit{image-to-pose} recognition model
 with the \textit{pose-to-image} generative model.
Even when only images without pose annotations are given,
 the generative model can be used
 for evaluating the anatomical fidelity of poses
 estimated by the recognition model
 (\ie, how consistent the estimated poses are with the given images).
In the same way,
 our method builds a \textit{pose-to-feature-to-pose} reflective model
 (\ie, a lower-level mirror system for pose understanding)
 by connecting the \textit{pose-to-feature} recognition model
 with \textit{feature-to-pose} generative model.
This pose VAE can be trained in advance by using a large number
 of pose data (\eg, those obtained by rendering human 3D models)
 and then used for evaluating the anatomical plausibility of the estimated poses.
Note that the pose VAE cannot be used alone
 as an evaluator of an estimated pose for a non-annotated image
 because \textit{any} plausible pose is allowed
 even if it does not reflect the image.
This is
 why conventional plausibility-aware methods still need paired data
 \cite{chen2017adversarial,chen2018cascaded,chou2018self}.
The higher- and lower- mirror systems are integrated into MirrorNet
 and can be trained jointly in a statistically principled manner.
In practice, each component of MirrorNet
 is trained separately by using paired data
 and then the entire MirrorNet is jointly trained
 in a semi-supervised manner using both paired and unpaired data for further optimization.

The main contributions of this paper are as follows.
We formulate a unified probabilistic model of poses and images,
 and propose a hierarchical autoencoding variational inference method
 based on a two-level mirror system
 for plausibility- and fidelity-aware pose estimation.
Our pose estimation method is the first
 that can use images with and without annotations for semi-supervised learning.
We experimentally show that the performance of the image-to-pose recognition model
 can be improved by integrating the pose-to-image generative model
 and the pose VAE.

The rest of this paper is organized as follows.
Section~\ref{sec:related_work} reviews related work
 on plausibility-aware pose estimation and fidelity-aware image processing.
Section~\ref{sec:proposed_method}
 explains the proposed method
 for unsupervised, supervised, and semi-supervised pose estimation.
Section~\ref{sec:implementation}
 describes the detailed implementation of the proposed method.
Section~\ref{sec:evaluation}
 reports comparative experiments conducted for evaluating the proposed method.
Section~\ref{sec:conclusion}
 summarizes this paper.

\section{Related Work}
\label{sec:related_work}

2D human pose estimation refers to estimating the coordinates of joints of a person in an image,
 in contrast to skeleton extraction~\cite{aubert2014poisson,whytock2014dynamic,shen2017deepskeleton}.
This task is challenging
 because a wide variety of human appearances and background scenes can exist and some joints are often occluded.

For robust pose estimation,
 Ramanan~\cite{ramanan2007learning} proposed an edge-based model,
 and Andriluka~\etal~\cite{andriluka2009pictorial}
 introduced a pictorial structure model of human joints.
Modeling the human body using tree or graph structures
 has been intensively studied~\cite{gkioxari2013articulated,sapp2010adaptive,yang2011articulated,johnson2011learning,sapp2013modec,pishchulin2013poselet,dantone2013human}.
To improve the accuracy of estimation,
 one needs to carefully design sophisticated models and features
 that can appropriately represent the relations between joints.

Toshev and Szegedy~\cite{toshev2014deeppose} proposed a neural pose estimator called DeepPose
 that estimates the positions of joints by using a DNN
 consisting of convolutional layers and fully connected layers.
DeepPose is the first method
 that applies deep learning to pose estimation,
 resulting in significant performance improvement.
Instead of directly regressing the coordinates of joints from an image as in DeepPose,
 Thompson~\etal~\cite{tompson2014joint} used a heatmap (pixel-wise likelihood)
 for representing the distributions of each joint,
 which has recently become standard.
The state-of-the-art methods of 2D pose estimation
 have been examined from several points of view.
For example, intermediate supervision and multi-stage learning
 were proposed for using the large receptive field of deep convolutional neural networks (CNNs)~\cite{newell2016stacked,wei2016convolutional,belagiannis2017recurrent,yang2017learning,sun2019deep}.
An optimal objective function was proposed for evaluating
 the relations between pairs of joints~\cite{lifshitz2016human,bulat2016human,chu2017multi,tang2018deeply}.
Recently, some studies have assessed
 the correctness of inferred poses using additional networks
 \cite{chen2018cascaded,fieraru2018learning,moon2019posefix}
 or compensated for the lack of data samples with data augmentation~\cite{ukita2018semi,peng2018jointly}.
We here review plausibility-aware methods of pose estimation
 and fidelity-aware methods of image processing.

\subsection{Plausibility-Aware Pose Estimation}

A standard way of improving the anatomical plausibility of estimated poses
 is to focus on the local relations of adjacent joints in pose estimation~\cite{lifshitz2016human,bulat2016human,chu2017multi,tang2018deeply,nie2018human}
 or to refine the estimated poses as post-processing.
Carreira~\etal~\cite{carreira2016human}
 proposed a self-correcting model based on iterative error feedback.
Chen~\etal~\cite{chen2018cascaded}, Fieraru~\etal~\cite{fieraru2018learning},
 and Moon~\etal~\cite{moon2019posefix} proposed cascaded networks
 that recursively refine the estimated poses while referring to the original images.
Adversarial networks have often been used
 to judge whether the estimated poses
 are anatomically plausible~\cite{chen2017adversarial,chou2018self}.
In addition,
Ke~\etal~\cite{ke2018multi} proposed a scale-robust method
 based on a multi-scale network with a body structure-aware loss function.
Nie~\etal~\cite{nie2019single} proposed a structured pose representation
 using the displacement in the position of every joint from a root joint position.
While these methods can use only paired data for supervised learning,
 our VAE-based method enables unsupervised learning.
In contract to the existing autoencoding approach
 that aims to extract latent features of poses~\cite{walker2017pose,ma2018disentangled},
 our VAE is used for measuring the plausibility of poses.

To compensate for lack of training data,
 Ukita and Uematsu~\cite{ukita2018semi} took a weakly supervised approach
 that uses action labels of images (\eg, baseball and volleyball)
 to estimate the poses of humans from a part of paired data.
Peng~\etal~\cite{peng2018jointly} proposed an efficient data augmentation method
 that generates hard-to-recognize images with adversarial training.
Yeh~\etal~\cite{yehchirality} used the chirality transform,
 a geometric transform that generates an antipode of a target,
 for pose regression.
In this paper, we take a different approach based on mirror systems for unsupervised learning
 such that non-annotated images can be used to improve the performance.

\subsection{Fidelity-Aware Image Processing}

Mirror structures have been used successfully
 for various image processing tasks including domain conversion.
Kingma and Welling~\cite{kingma2014autoencoding}
 proposed the VAE
 that jointly learns a generative model (decoder) of observed variables
 from latent variables following a prior distribution,
 and a recognition model (encoder) of the latent variables from the observed variables.
It can generate new samples
 by randomly drawing latent variables from the prior distribution.
CycleGAN~\cite{zhu2017unpaired}, DiscoGAN~\cite{kim2017learning},
 and DualGAN~\cite{yi2017dualgan} are popular variants of GANs
 using mirror structures for image-to-image conversion.
The key feature of these methods is to consider mutual mappings
 between domains from unpaired data.
Qiao~\etal~\cite{qiao2019mirrorgan} recently proposed MirrorGAN
 for bidirectional inter-domain (text-image) conversion.
Yildiri~\etal~\cite{Yildirimeaax5979} proposed an analysis-by-synthesis approach
 to joint 3D face generation and recognition from a cognitive point of view.
The success of these methods
 indicates the potential of mirror structures
 for stably training a DNN with unsupervised data.
In this paper, we propose
 the first mirror-structured DNN for human pose estimation
 that integrates two-level mirror systems
 in a hierarchically autoencoding variational manner.


\section{Proposed Method}
\label{sec:proposed_method}

\begin{figure*}[!tb]
\centering
\includegraphics[width=.86\linewidth]{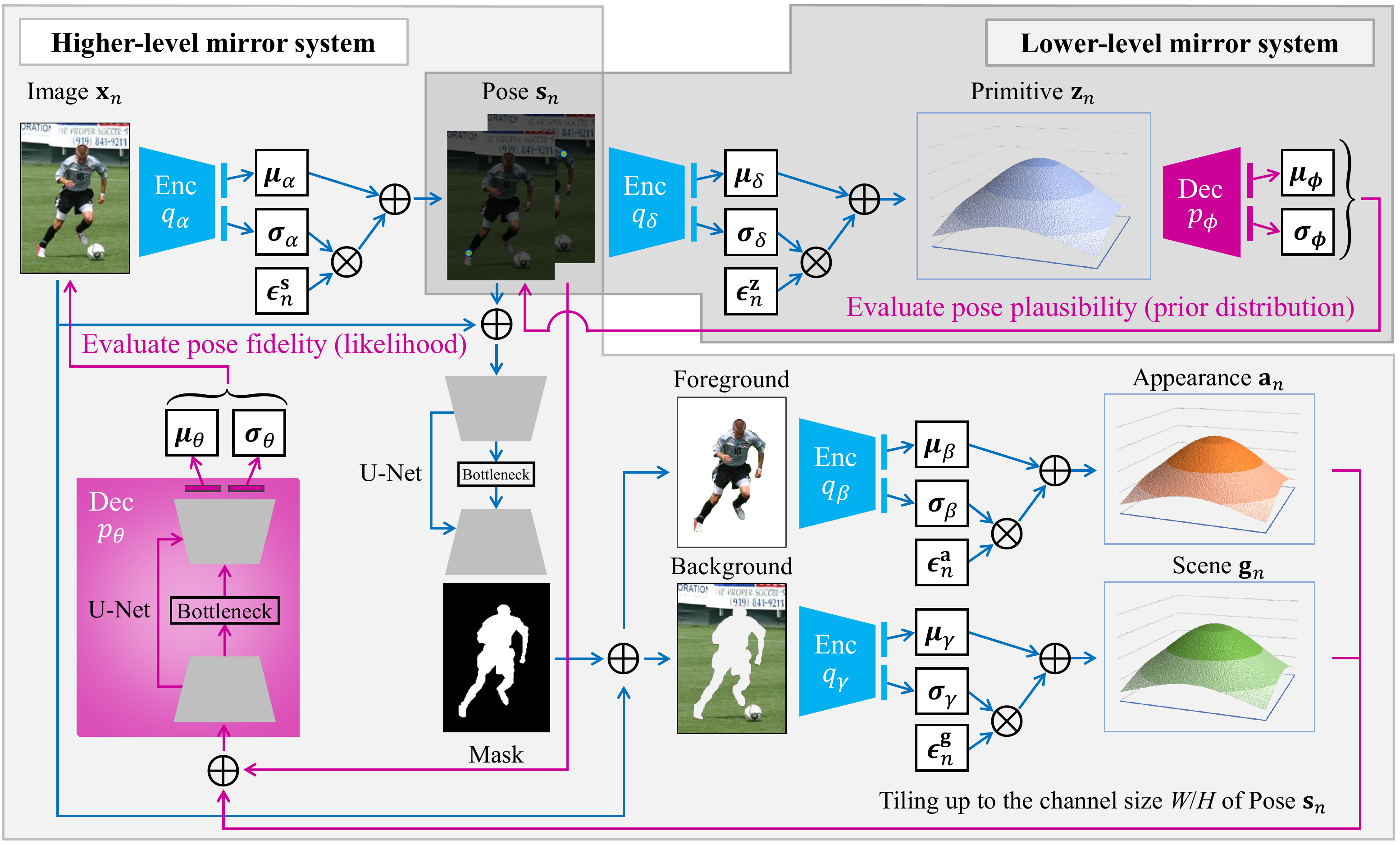}
\caption{
The proposed architecture of MirrorNet integrating a VAE for poses (lower-level mirror system)
 with a pose-conditioned VAE for images (higher-level mirror system)
 in a hierarchical Bayesian manner.
In terms of generative modeling,
 the decoder of the pose VAE serves 
 as a prior distribution of poses $p(\mb{S})$ to evaluate the pose plausibility 
 and the decoder of the image VAE 
 as a likelihood function of poses $p(\mb{X}|\mb{S})$ to evaluate the pose fidelity.
In terms of posterior inference,
 the encoder of the pose VAE is used as a variational posterior distribution of poses $q(\mb{S}|\mb{X})$.
Such a statistical approach based on a complete probabilistic generative model
 enables \textit{semi-supervised} pose estimation using any images with/without pose annotations
}
\label{fig:architecture}
\end{figure*}
This section describes the proposed method
 based on a fully probabilistic model of poses and images
 for 2D pose estimation in images of people (\fref{fig:architecture}).
MirrorNet is a hierarchical VAE,
 one of the techniques of amortized variational inference (AVI)~\cite{kingma2014autoencoding,dai2016variational,mnih2014neural,rezende2014stochastic},
 and consists of a VAE of images (\ie, a \textit{pose-to-image} generative model
 and an \textit{image-to-pose} recognition model),
 and a VAE of poses (\ie, a \textit{primitive-to-pose} generative model
 and a \textit{pose-to-primitive} recognition model).
In theory, this model can be trained
 in an unsupervised manner by using non-annotated images only,
 or by using unpaired images and poses.
In practice, the model is trained in a semi-supervised manner
 by using partially annotated images.
Each model is first trained separately to stabilize the training,
 and then all models are jointly trained for further optimization.
Once the training is completed,
 only the image-to-pose recognition model is used for pose estimation.
The hierarchical autoencoding architecture is effective
 for estimating poses that are anatomically plausible
 and have high fidelity to the original images.

\subsection{Problem Specification}
Let $\mb{X} = \{\mb{x}_n \in \mathbb{R}^{D^\mb{x}}\}_{n=1}^N$
 and $\mb{S} = \{\mb{s}_n \in \mathbb{R}^{D^\mb{s}}\}_{n=1}^N$
 be a set of images
 and a set of poses corresponding to $\mb{X}$, respectively,
 where $D^\mb{x} (\sim\mathcal{O}(10^6))$ is the number of dimensions of each image, $D^\mb{s} (\sim\mathcal{O}(10^5))$ is the number of dimensions of each pose,
 and $N$ is the number of images.
We assume that each $\mb{x}_n$ is an RGB image featuring
 a single or multiple people, showing all or parts of their bodies,
 and each $\mb{s}_n$ is a set of grayscale images,
 each of which represents the position of a joint using a heatmap~\cite{tompson2014joint}.

Let $\mb{A} = \{\mb{a}_n \in \mathbb{R}^{D^\mb{a}}\}_{n=1}^N$
 and $\mb{G} = \{\mb{g}_n \in \mathbb{R}^{D^\mb{g}}\}_{n=1}^N$
 be a set of \textit{appearances}
 representing the foreground features of $\mb{X}$
 (\eg, skin and hair colors and textures)
 and a set of \textit{scenes}
 representing the background features of $\mb{X}$
 (\eg, places, color, and brightness), respectively,
 where $D^\mb{a}$ and $D^\mb{g}$ are the number of dimensions of the latent spaces.
These latent features
 are used in combination with $\mb{S}$ for representing $\mb{X}$.
Let $\mb{Z} = \{\mb{z}_n \in \mathbb{R}^{D^\mb{z}}\}_{n=1}^N$
 be a set of \textit{primitives}
 representing the features of $\mb{S}$
 (\eg, scales, positions, and orientations of joints),
 where $D^\mb{z}$ is the number of dimensions of the latent space.

Our goal is to train a pose estimator
 that maps $\mb{X}$ to $\mb{S}$.
Let $M$ be the number of annotated images.
In a supervised condition,
 $\mb{X}$ and $\mb{S}$ are given as observed data ($M=N$).
In an unsupervised condition,
 only $\mb{X}$ is given ($M=0$).
In a semi-supervised condition,
 $\mb{X}$ and a part of $\mb{S}$, \ie, $\{\mb{s}_n\}_{n=1}^M$, are given.

\begin{figure}[!tb]
\centering
\includegraphics[width=.86\linewidth]{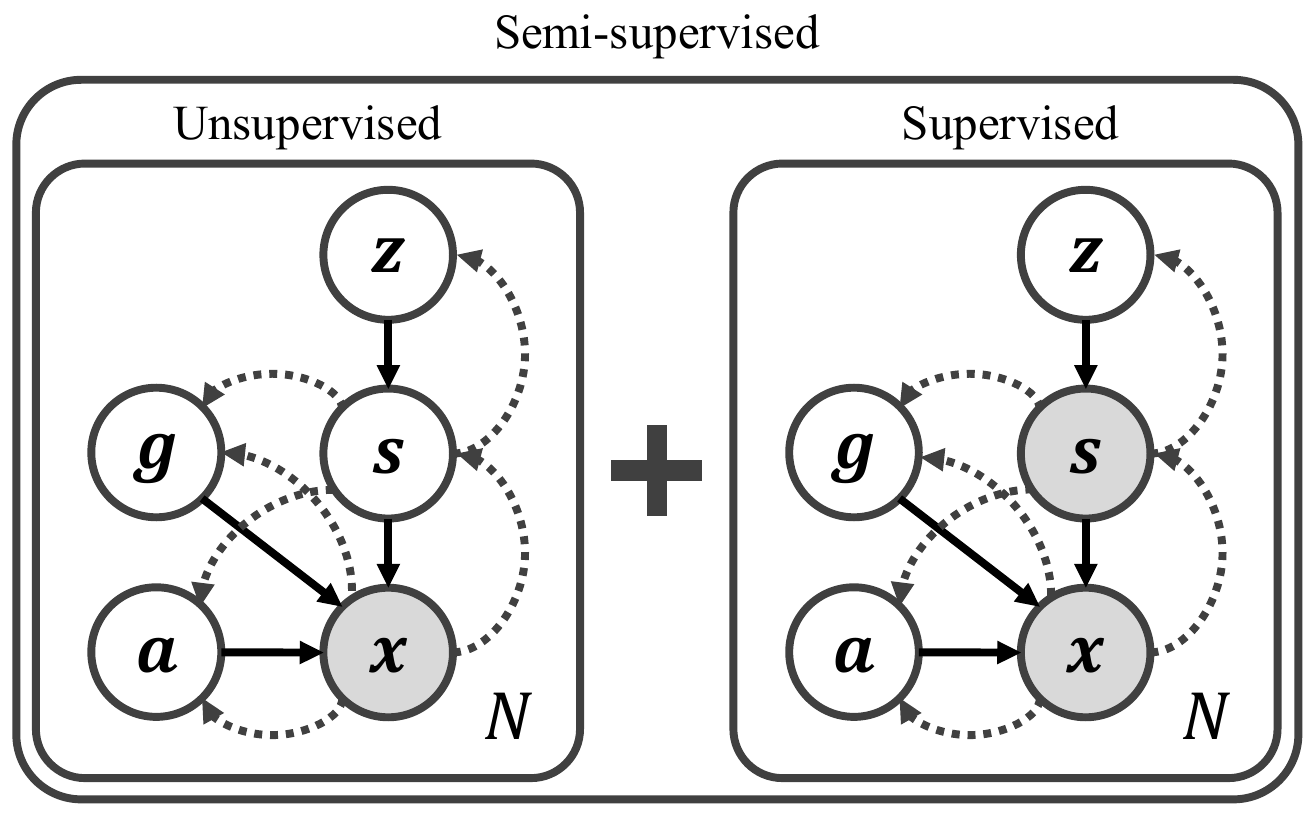}
\caption{
Three types of statistical inference. 
The solid and dashed arrows indicate the generative and inference models, respectively,
 and the shaded circles represent the given data.
In the unsupervised learning (Section \ref{sec:unsupervised_learning}), 
 the networks are trained using only images $\mb{X}$.
In the supervised learning (Section \ref{sec:supervised_learning}), 
 the networks are trained using paired data of images $\mb{X}$ with poses $\mb{S}$.
The semi-supervised learning is a mixture of these two conditions
}
\label{fig:graph}
\end{figure}

\subsection{Generative Modeling}
We formulate a unified hierarchical generative model
 of images $\mb{X}$, poses $\mb{S}$, appearances $\mb{A}$, scenes $\mb{G}$, and primitives $\mb{Z}$
 that integrates a deep generative model of $\mb{X}$ from $\mb{S}$, $\mb{A}$, and $\mb{G}$
 with a deep generative model of $\mb{S}$ from $\mb{Z}$
 as follows (Fig.~\ref{fig:graph}):
\begin{align}
&\!
p(\mb{X},\mb{S},\mb{A},\mb{G},\mb{Z})
\nonumber\\
&\!
=
p_\theta(\mb{X}|\mb{S},\mb{A},\mb{G}) p_\phi(\mb{S}|\mb{Z}) p(\mb{A}) p(\mb{G}) p(\mb{Z})
\nonumber\\
&\!
=
\prod_{n=1}^N p_\theta(\mb{x}_n|\mb{s}_n,\mb{a}_n,\mb{g}_n) p_\phi(\mb{s}_n|\mb{z}_n)p(\mb{a}_n)p(\mb{g}_n) p(\mb{z}_n),
\!
\label{eq:joint_dist}
\end{align}
where $\theta$ and $\phi$
 are the sets of trainable parameters of the deep generative models of $\mb{x}_n$ and $\mb{s}_n$, respectively.
The pose likelihood $p_\theta(\mb{x}_n|\mb{s}_n,\mb{a}_n,\mb{g}_n)$
 evaluates the pose fidelity to the given images
 and the pose prior $p_\phi(\mb{s}_n|\mb{z}_n)$
 prevents anatomically implausible pose estimates.
The remaining terms
 are priors of $\mb{a}_n,\mb{g}_n$, and $\mb{z}_n$.

The \textit{pose-to-image} generation model
 $p_\theta(\mb{x}_n|\mb{s}_n,\mb{a}_n,\mb{g}_n)$
 and the \textit{primitive-to-pose} generation model
 $p_\phi(\mb{s}_n|\mb{z}_n)$ are both formulated as follows:
\begin{align}
 &p_\theta(\mb{x}_n|\mb{s}_n,\mb{a}_n,\mb{g}_n) \nonumber\\
 &= \mathcal{N}(\mb{x}_n;\mu_\theta(\mb{s}_n,\mb{a}_n,\mb{g}_n),
 \sigma_\theta^2(\mb{s}_n,\mb{a}_n,\mb{g}_n) \mb{I}_{D^\mb{x}}),
 \label{eq:p_x}
 \\
 &p_\phi(\mb{s}_n|\mb{z}_n)
 = \mathcal{N}(\mb{s}_n;\mu_\phi(\mb{z}_n), \sigma_\phi^2(\mb{z}_n) \mb{I}_{D^\mb{s}}),
 \label{eq:p_s}
\end{align}
where $\mb\mu_\theta(\mb{s}_n,\mb{a}_n,\mb{g}_n)$
 and $\sigma_\theta(\mb{s}_n,\mb{a}_n,\mb{g}_n)$ are the outputs
 of a DNN with parameter $\theta$
 that takes $\mb{s}_n, \mb{a}_n$, and $\mb{g}_n$ as input,
 and $\mb\mu_\phi(\mb{z}_n)$ and $\sigma_\phi(\mb{z}_n)$ are the outputs
 of a DNN with parameters $\phi$ that takes $\mb{z}_n$ as input.
The priors $p(\mb{a}_n)$, $p(\mb{g}_n)$, and $p(\mb{z}_n)$ are set to the standard Gaussian distributions as follows:
\begin{align}
 p(\mb{a}_n) &= \mathcal{N}(\mb{a}_n ; \zeros_{D^\mb{a}}, \mb{I}_{D^\mb{a}}),
 \label{eq:p_a}
 \\
 p(\mb{g}_n) &= \mathcal{N}(\mb{g}_n ; \zeros_{D^\mb{g}}, \mb{I}_{D^\mb{g}}),
 \label{eq:p_g}
 \\
 p(\mb{z}_n) &= \mathcal{N}(\mb{z}_n ; \zeros_{D^\mb{z}}, \mb{I}_{D^\mb{z}}),
 \label{eq:p_z}
\end{align}
where $\zeros_{D^\dag}$ $(D^\dag = \{D^\mb{a}, D^\mb{g}, D^\mb{z}\})$ and $\mb{I}_{D^\dag}$ are
 the zero vector of size $D^\dag$ and the identity matrix of size $D^\dag$, respectively.

\subsection{Unsupervised Learning}
\label{sec:unsupervised_learning}

We explain the unsupervised learning of the proposed model using only images $\mb{X}$,
 which is the basis for practical semi-supervised learning
 using \textit{partially} annotated images (Section \ref{sec:supervised_learning}).
Given a set of images $\mb{X}$ as observed data,
 our goal is to infer the distribution of the latent variables
 $\mb\Omega \equiv (\mb{S}, \mb{A}, \mb{G}, \mb{Z})$.
We estimate optimal parameters $\theta^*$ and $\phi^*$
 in the framework of maximum likelihood estimation as follows:
\begin{align}
 \theta^*, \phi^*
 =
 \Argmax_{\theta,\phi} p(\mb{X}),
\end{align}
and $p(\mb{X})$ is the marginal likelihood given by
\begin{align}
 p(\mb{X}) = \int\! p(\mb{X}, \mb\Omega) d \mb\Omega.
 \label{eq:p_mar_dist}
\end{align}
where $p(\mb{X},\mb\Omega)$ is the joint probability distribution given by \eref{eq:joint_dist}.

Because \eref{eq:p_mar_dist} is analytically intractable,
 we use an amortized variational inference (AVI) technique~\cite{kingma2014autoencoding,dai2016variational,mnih2014neural,rezende2014stochastic}
 that introduces an arbitrary variational posterior distribution $q(\mb\Omega|\mb{X})$
 and makes it as close to the true posterior distribution $p(\mb\Omega|\mb{X})$
 (Section \ref{sec:variational_lower_bound}).
The minimization of the Kullback--Leibler (KL) divergence between these posteriors
 is equivalent to the maximization
 of a variational lower bound $\mathcal{L}$ of $\log p(\mb{X})$ with respect to $q(\mb\Omega|\mb{X})$.
Thus, the optimal parameters $\theta^*$ and $\phi^*$ can be obtained
 by maximizing the variational lower bound $\mathcal{L}$ instead of $\log p(\mb{X})$
 (Section \ref{sec:parameter_optimization}).

\subsubsection{Variational Lower Bound}
\label{sec:variational_lower_bound}

Using Jensen's inequality,
 a variational lower bound $\mathcal{L}^\mb{X}$ of $\log p(\mb{X})$
 can be derived as follows:
\begin{align}
 \log p(\mb{X})
 &=
 \log \int\! p(\mb{X},\mb\Omega) d\mb\Omega
 \nonumber\\
 &=
 \log \int\! \frac{q(\mb\Omega|\mb{X})}{q(\mb\Omega|\mb{X})} p(\mb{X},\mb\Omega) d\mb\Omega
 \nonumber\\
 &\geq
 \int\!  q(\mb\Omega|\mb{X})\log \frac{p(\mb{X},\mb\Omega)}{q(\mb\Omega|\mb{X})} d\mb\Omega
 \overset{\mbox{\tiny def}}{=}
 \mathcal{L}^\mb{X},
 \label{eq:lower_bound}
\end{align}
where the equality holds, \ie, $\mathcal{L}^\mb{X}$ is maximized,
 if and only if $q(\mb\Omega|\mb{X}) = p(\mb\Omega|\mb{X})$.
Because this equality condition cannot be computed analytically,
 $q(\mb\Omega|\mb{X})$ is approximated by a factorized form
 as follows:
\begin{align}
 &q(\mb\Omega|\mb{X})
 =
 q_\alpha(\mb{S}|\mb{X}) q_\beta(\mb{A}|\mb{S},\mb{X})
 q_\gamma(\mb{G}|\mb{S},\mb{X}) q_\delta(\mb{Z}|\mb{S})
 \nonumber\\
 &\!=
 \prod_{n=1}^N q_\alpha(\mb{s}_n|\mb{x}_n) q_\beta(\mb{a}_n|\mb{s}_n,\mb{x}_n)
 q_\gamma(\mb{g}_n|\mb{s}_n,\mb{x}_n) q_\delta(\mb{z}_n|\mb{s}_n),
 \label{eq:var_dist}
\end{align}
where $\alpha$, $\beta$, $\gamma$, and $\delta$ are the sets of parameters
 of these four variational distributions, respectively.

In the statistical framework of AVI,
 we introduce a DNN-based posterior distribution $q(\mb\Omega|\mb{X})$
 such that the complex true posterior distribution $p(\mb\Omega|\mb{X})$
 can be well approximated by $q(\mb\Omega|\mb{X})$.
Specifically,
 we introduce
 a deep \textit{image-to-pose} model $q_\alpha(\mb{s}_n|\mb{x}_n)$,
 a deep \textit{image-to-appearance} model $q_\beta(\mb{a}_n|\mb{s}_n,\mb{x}_n)$,
 a deep \textit{image-to-scene} model $q_\gamma(\mb{g}_n|\mb{s}_n,\mb{x}_n)$,
 and a deep \textit{pose-to-primitive} model $q_\delta(\mb{z}_n|\mb{s}_n)$ as follows:
\begin{align}
 q_\alpha(\mb{s}_n|\mb{x}_n)
 &= \mathcal{N}(\mb{s}_n;\mb\mu_\alpha(\mb{x}_n), \sigma_\alpha^2(\mb{x}_n)\mb{I}_{D^\mb{s}}),
 \label{eq:q_s}
 \\
 q_\beta(\mb{a}_n|\mb{s}_n,\mb{x}_n)
 &= \mathcal{N}(\mb{a}_n;\mb\mu_\beta(\mb{s}_n,\mb{x}_n), \mb\sigma_\beta^2(\mb{s}_n,\mb{x}_n)\mb{I}_{D^\mb{a}}),\!
 \label{eq:q_a}
 \\
 q_\gamma(\mb{g}_n|\mb{s}_n,\mb{x}_n)
 &= \mathcal{N}(\mb{g}_n;\mb\mu_\gamma(\mb{s}_n,\mb{x}_n), \mb\sigma_\gamma^2(\mb{s}_n,\mb{x}_n)\mb{I}_{D^\mb{g}}),\!
 \label{eq:q_g}
 \\
 q_\delta(\mb{z}_n|\mb{s}_n)
 &= \mathcal{N}(\mb{z}_n;\mb\mu_\delta(\mb{s}_n), \mb\sigma_\delta^2(\mb{s}_n)\mb{I}_{D^\mb{z}}),
 \label{eq:q_z}
\end{align}
where $\mb\mu_\alpha(\mb{x}_n)$ and $\sigma_\alpha(\mb{x}_n)$ are the outputs
 of a DNN with parameters $\alpha$
 that takes $\mb{x}_n$ as input,
 $\mb\mu_{\ddag}(\mb{s}_n,\mb{x}_n)$
 and $\mb\sigma_{\ddag}(\mb{s}_n,\mb{x}_n)$ $(\ddag = \beta \ \mathrm{or} \ \gamma)$ are the outputs
 of a DNN with parameters $\ddag$
 that takes $\mb{s}_n$ and $\mb{x}_n$ as input,
 and $\mb\mu_\delta(\mb{s}_n)$ and $\mb\sigma_\delta(\mb{s}_n)$ are the outputs
 of a DNN with parameters $\delta$
 that takes $\mb{s}_n$ as input.

Substituting both of the generative model given by \eref{eq:joint_dist}
 with Eqs.~(\ref{eq:p_x})--(\ref{eq:p_z})
 and the recognition model given by  \eref{eq:var_dist}
 with Eqs.~(\ref{eq:q_s})--(\ref{eq:q_z})
 into \eref{eq:lower_bound},
 the variational lower bound $\mathcal{L}^\mb{X}$ can be rewritten
 as the sum of $\{\mathcal{L}^\mb{X}_n\}_{n=1}^N$ ($\mathcal{L}^\mb{X} = \sum_n \mathcal{L}^\mb{X}_n$)
 as follows (Appendix \ref{sec:lowerbound}):
\begin{align}
 \mathcal{L}^\mb{X}_n
 =&
 \ \mathbb{E}_q
 [\log p_\theta(\mb{x}_n|\mb{s}_n,\mb{a}_n,\mb{g}_n) + \log p_\phi(\mb{s}_n|\mb{z}_n)
 \nonumber\\
 &
 + \log p(\mb{a}_n) + \log p(\mb{g}_n) + \log p(\mb{z}_n)
 \nonumber\\
 &
 - \log q_\alpha(\mb{s}_n|\mb{x}_n) - \log q_\beta(\mb{a}_n|\mb{s}_n,\mb{x}_n)
 \nonumber\\
 &
 - \log q_\gamma(\mb{g}_n|\mb{s}_n,\mb{x}_n) - \log q_\delta(\mb{z}_n|\mb{s}_n)]
 \nonumber\\
 =&
 \ \mathbb{E}_q [\log p_\theta(\mb{x}_n|\mb{s}_n,\mb{a}_n,\mb{g}_n)]
 \nonumber\\
 &
 + \mathbb{E}_q [\log p_\phi(\mb{s}_n|\mb{z}_n)]
 \nonumber\\
 &
 - \mathbb{E}_q [\log q_\alpha(\mb{s}_n|\mb{x}_n)]
 \nonumber\\
 &
 - \mathbb{E}_q [\mbox{KL}(q_\beta(\mb{a}_n|\mb{s}_n,\mb{x}_n) || p(\mb{a}_n))]
 \nonumber\\
 &
 - \mathbb{E}_q [\mbox{KL}(q_\gamma(\mb{g}_n|\mb{s}_n,\mb{x}_n) || p(\mb{g}_n))]
 \nonumber\\
 &
 - \mathbb{E}_q [\mbox{KL}(q_\delta(\mb{z}_n|\mb{s}_n) || p(\mb{z}_n))],
 \label{eq:derivation_lower_bound}
\end{align}
where
 the first term represents the fidelity of a pose $\mb{s}_n$
 with an original image $\mb{x}_n$ having features $\mb{a}_n$ and $\mb{g}_n$,
 the second term represents the plausibility of $\mb{s}_n$,
 the third term prevents the overfitting of the recognition model $\alpha$,
 and the fourth to sixth terms evaluate the similarities
 between the recognition models $\beta$, $\gamma$, and $\delta$
 and the priors on $\mb{a}_n$, $\mb{g}_n$, and $\mb{z}_n$, respectively.

\subsubsection{Parameter Optimization}
\label{sec:parameter_optimization}

Because \eref{eq:derivation_lower_bound} still includes intractable expectations,
 we perform Monte Carlo integration
 using samples $\mb{s}_n$, $\mb{a}_n$, $\mb{g}_n$, and $\mb{z}_n$
 obtained by \textit{reparametrization trick}~\cite{kingma2014autoencoding} as follows:
\begin{align}
 \mb\epsilon^\mb{s}_n &\sim \mathcal{N}(\zeros, \mb{I}_{D^\mb{s}}),
 \\
 \mb\epsilon^\mb{a}_n &\sim \mathcal{N}(\zeros, \mb{I}_{D^\mb{a}}),
 \\
 \mb\epsilon^\mb{g}_n &\sim \mathcal{N}(\zeros, \mb{I}_{D^\mb{g}}),
 \\
 \mb\epsilon^\mb{z}_n &\sim \mathcal{N}(\zeros, \mb{I}_{D^\mb{z}}),
 \\
 \mb{s}_n &= \mb\mu_\alpha(\mb{x}_n) + \mb\epsilon^\mb{s}_n \odot \mb\sigma_\alpha(\mb{x}_n),
 \label{eq:sampling_sn}
 \\
 \mb{a}_n &= \mb\mu_\beta(\mb{s}_n,\mb{x}_n) + \mb\epsilon^\mb{a}_n \odot \mb\sigma_\beta(\mb{s}_n,\mb{x}_n),
 \label{eq:sampling_an}
 \\
 \mb{g}_n &= \mb\mu_\gamma(\mb{s}_n,\mb{x}_n) + \mb\epsilon^\mb{g}_n \odot \mb\sigma_\gamma(\mb{s}_n,\mb{x}_n),
 \label{eq:sampling_gn}
 \\
 \mb{z}_n &= \mb\mu_\delta(\mb{s}_n) + \mb\epsilon^\mb{z}_n \odot \mb\sigma_\delta(\mb{s}_n),
 \label{eq:sampling_zn}
\end{align}
where $\odot$ indicates the element-wise product.
Although in theory a sufficient number of samples
 should be generated to perform accurate Monte Carlo integration,
 we generate only one sample for each variable as in the standard VAE~\cite{kingma2014autoencoding}.

Using these tricks,
 the lower bound $\mathcal{L}^\mb{X}$
 given by \eref{eq:derivation_lower_bound} can be approximately computed,
 and can thus be maximized with respect to $\theta$, $\phi$, $\alpha$, $\beta$, $\gamma$, and $\delta$
 (Fig.~\ref{fig:architecture}).
First,
 the recognition models $\alpha$, $\beta$, $\gamma$, and $\delta$
 are used to \textit{deterministically} generate samples $\mb{s}_n$, $\mb{a}_n$, $\mb{g}_n$, and $\mb{z}_n$
 in Eqs.~(\ref{eq:sampling_sn})--(\ref{eq:sampling_zn}),
 and to calculate the last four regularization terms of \eref{eq:derivation_lower_bound}, respectively.
Given the samples $\mb{s}_n$, $\mb{a}_n$, $\mb{g}_n$, and $\mb{z}_n$,
 the generative models $\theta$ and $\phi$ are used
 to calculate the first two reconstruction terms of \eref{eq:derivation_lower_bound}, respectively.
The recognition models $\alpha$, $\beta$, $\gamma$, and $\delta$,
 and the generative models $\phi$ and $\theta$
 can thus be concatenated in this order
 with the reparametrization trick given
 by Eqs.~(\ref{eq:sampling_sn})--(\ref{eq:sampling_zn}),
 and are jointly optimized in an autoencoding manner
 with an objective function given by \eref{eq:derivation_lower_bound}.

\subsection{Supervised Learning}
\label{sec:supervised_learning}

We explain the supervised learning of the proposed model
 using paired data of $\mb{X}$ and $\mb{S}$.
This approach follows the manner of the semi-supervised learning of a VAE~\cite{kingma2014semisupervised}.
While the variational lower bound $\mathcal{L}^\mb{X}$ of $\log p(\mb{X})$
 is maximized in the unsupervised condition (Section \ref{sec:unsupervised_learning}),
 we aim to maximize the variational lower bound $\mathcal{L}^{\mb{X},\mb{S}}$ of $\log p(\mb{X},\mb{S})$,
 which is given by
\begin{align}
 \log p(\mb{X},\mb{S})
 &=
 \log \! \int\! p(\mb{X},\mb\Omega) d\mb\Theta
 \nonumber\\
 &
 =
 \log \! \int\! \frac{q(\mb\Theta|\mb{S},\mb{X})}{q(\mb\Theta|\mb{S},\mb{X})} p(\mb{X},\mb\Omega) d\mb\Theta
 \nonumber\\
 &\geq
 \int\! q(\mb\Theta|\mb{S},\mb{X})\log \frac{p(\mb{X},\mb\Omega)}{q(\mb\Theta|\mb{S},\mb{X})} d\mb\Theta
 \nonumber\\
 &\overset{\mbox{\tiny def}}{=}
 \mathcal{L}^{\mb{X},\mb{S}},
 \label{eq:lower_bound_semi}
\end{align}
where $\mb\Theta = \mb\Omega \! \setminus \! \mb{S} = \{\mb{A}, \mb{G}, \mb{Z}\}$.
As in \eref{eq:var_dist},
 $q(\mb\Theta|\mb{S},\mb{X})$ is factorized as
\begin{align}
 &q(\mb\Theta|\mb{S},\mb{X})
 \nonumber\\
 &=
 q_\beta(\mb{A}|\mb{S},\mb{X}) q_\gamma(\mb{G}|\mb{S},\mb{X}) q_\delta(\mb{Z}|\mb{S})
 \nonumber\\
 &=
 \prod_{n=1}^N q_\beta(\mb{a}_n|\mb{s}_n,\mb{x}_n) q_\gamma(\mb{g}_n|\mb{s}_n,\mb{x}_n) q_\delta(\mb{z}_n|\mb{s}_n),
 \label{eq:var_dist_sv}
\end{align}
where $q_\beta(\mb{a}_n|\mb{s}_n,\mb{x}_n)$, $q_\gamma(\mb{g}_n|\mb{s}_n,\mb{x}_n)$, and $q_\delta(\mb{z}_n|\mb{s}_n)$
 are given by Eqs.~(\ref{eq:q_a})--(\ref{eq:q_z}), respectively.
Substituting both of the probabilistic model
 given by \eref{eq:joint_dist} with Eqs.~(\ref{eq:p_x})--(\ref{eq:p_z})
 and the inference model given by \eref{eq:var_dist_sv} with Eqs.~(\ref{eq:q_a})--(\ref{eq:q_z})
 into \eref{eq:lower_bound_semi},
 $\mathcal{L}^{\mb{X},\mb{S}}$ can be rewritten
 as the sum of $\{\mathcal{L}^{\mb{X},\mb{S}}_n\}_{n=1}^N$
 ($\mathcal{L}^{\mb{X},\mb{S}} = \sum_n \mathcal{L}^{\mb{X},\mb{S}}_n$)
 as follows (Appendix \ref{sec:lowerbound}):
\begin{align}
 \mathcal{L}^{\mb{X},\mb{S}}_n
 =&
 \ \mathbb{E}_q
 [
 \log p_\theta(\mb{x}_n|\mb{s}_n,\mb{a}_n,\mb{g}_n) + \log p_\phi(\mb{s}_n|\mb{z}_n)
 \nonumber\\
 &
 + \log p(\mb{a}_n) + \log p(\mb{g}_n) + \log p(\mb{z}_n)
 \nonumber\\
 &
 - \log q_\beta(\mb{a}_n|\mb{s}_n,\mb{x}_n) - \log q_\gamma(\mb{g}_n|\mb{s}_n,\mb{x}_n)
 \nonumber\\
 &
 - \log q_\delta(\mb{z}_n|\mb{s}_n)]
 \nonumber\\
 =&
 \ \mathbb{E}_q
 [\log p_\theta(\mb{x}_n|\mb{s}_n,\mb{a}_n,\mb{g}_n)]
 \nonumber\\
 &
 + \mathbb{E}_q
 [\log p_\phi(\mb{s}_n|\mb{z}_n)]
 \nonumber\\
 &
 - \mbox{KL}(q_\beta(\mb{a}_n|\mb{s}_n,\mb{x}_n) || p(\mb{a}_n))
 \nonumber\\
 &
 - \mbox{KL}(q_\gamma(\mb{g}_n|\mb{s}_n,\mb{x}_n) || p(\mb{g}_n))
 \nonumber\\
 &
 - \mbox{KL}(q_\delta(\mb{z}_n|\mb{s}_n) || p(\mb{z}_n)).
 \label{eq:derivation_lower_bound_supervised}
\end{align}
A major problem in such supervised learning
 is that the recognition model $q_\alpha(\mb{s}_n|\mb{x}_n)$,
 which plays a central role for human pose estimation from images,
 cannot be trained
 because it does not appear in \eref{eq:derivation_lower_bound_supervised}.
To solve this problem,
 we add a term to assess the predictive performance of $q_\alpha(\mb{s}_n|\mb{x}_n)$
 to $\mathcal{L}_n$, following~\cite{kingma2014semisupervised} as
\begin{align}
 &\mathcal{L}^{\mb{X},\mb{S}}_{n,\lambda}
 = \mathcal{L}^{\mb{X},\mb{S}}_n + \lambda \log q_\alpha(\mb{s}_n|\mb{x}_n),
 \label{eq:l_n_lambda}
 \\
 &\log q_\alpha(\mb{s}_n|\mb{x}_n)
 \nonumber\\
 &=
 -\frac{1}{2} \sum^{D^\mb{s}}_{d_s=1}
 \left(\log \left(2\pi \sigma^2_{\alpha, d_s}(\mb{x}_n) \right)
 + \frac{(\mb{s}_{n} - \mu_{\alpha, d_s}(\mb{x}_n))^2}{\sigma_{\alpha, d_s}^2(\mb{x}_n)}\right),
\end{align}
where $\lambda$ is a hyperparameter that controls the balance
 between purely generative learning and purely discriminative learning.
In our method, we used $\lambda = 0.01$ in all experiments.
The new objective function $\mathcal{L}^{\mb{X},\mb{S}}_\lambda = \sum_n \mathcal{L}^{\mb{X},\mb{S}}_{n,\lambda}$
 can be maximized with respect to $\theta$, $\phi$, $\alpha$, $\beta$, $\gamma$, and $\delta$
 jointly in the same way as the unsupervised learning
 described in Section~\ref{sec:parameter_optimization},
 where $\mb{a}_n$, $\mb{g}_n$, and $\mb{z}_n$ are obtained by using Eqs. (\ref{eq:sampling_an})--(\ref{eq:sampling_zn}),
 and $\mb{s}_n$ is given.

\subsection{Semi-supervised Learning}
\label{sec:semi-supervised_learning}

In the semi-supervised condition, where $\mb{X}$ is only partially annotated,
 we define a new objective function $\mathcal{L}$
 by accumulating $\mathcal{L}^\mb{X}_n$ used for unsupervised learning
 or $\mathcal{L}^{\mb{X},\mb{S}}_{n,\lambda}$ used for supervised learning as follows:
\begin{align}
 \mathcal{L}
 \overset{\mbox{\tiny def}}{=}
 \sum_{n: \ \mbox{\scriptsize $\mb{x}_n$ is given}} \mathcal{L}^\mb{X}_n
 + \sum_{n: \ \mbox{\scriptsize $\mb{x}_n$ \& $\mb{s}_n$ are given}} \mathcal{L}^{\mb{X},\mb{S}}_{n,\lambda}.
\label{eq:lowerbound_sum}
\end{align}
All generation and recognition models
 can be trained for all samples regardless of the availability of their annotations.
In practice, it is effective to pre-train each model in advance.


\section{Implementation}
\label{sec:implementation}

This section describes the implementation of MirrorNet, which is
 based on curriculum learning.
First,
 we separately pre-train the components of MirrorNet,
 \ie, the pose recognizer $\alpha$ (\sref{sec:pose_recognizer}),
 the pose-conditioned image VAE
 with the generator $\theta$ and the recognizers $\beta$ and $\gamma$ (\sref{sec:image_vae}),
 and the pose VAE
 with the generator $\phi$ and the recognizer $\delta$ (\sref{sec:pose_vae}).
We then train the whole MirrorNet
 under a supervised condition (\sref{sec:supervised_learning})
 and further optimize it
 under a semi-supervised condition (\sref{sec:semi-supervised_learning}).

Note that, as shown in Fig.~\ref{fig:architecture}, the pose-conditioned image VAE has a human mask estimator as a subcomponent
 for separating an image into foreground and background images;
 this helps to stabilize the training of MirrorNet.

\subsection{Pose Recognizer}
\label{sec:pose_recognizer}
The image-to-pose recognizer $\alpha$ is of most interest in pose estimation,
 and is pre-trained in a \textit{supervised} manner
 by using paired data of $\mb{X}$ and $\mb{S}$.
We maximize an objective function given by
\begin{align}
 \mathcal{L}_n(\alpha)
 \overset{\mbox{\tiny def}}{=}
 \log q_\alpha(\mb{s}_n|\mb{x}_n),
 \label{eq:l_a}
\end{align}
where the variance $\sigma_\alpha^2(\mb{x}_n)$ is fixed to 0.01 for stability.

The network $\alpha$ can be implemented with any DNN
 that outputs the heatmaps of joint positions.
For this part, our implementation has three variations:
 a stack of eight residual hourglasses~\cite{newell2016stacked},
 ResNet-50 (baseline)~\cite{xiao2018simple},
 and high-resolution subnetworks~\cite{sun2019deep}.

\subsection{Pose-conditioned Image VAE}
\label{sec:image_vae}

The pose-conditioned image VAE
 consisting of the image-to-appearance recognizer $\beta$,
 the image-to-scene recognizer $\gamma$ (encoders),
 and the appearance/scene-to-image generator $\theta$ (decoder)
 is pre-trained in an \textit{unsupervised} manner
 by using paired data of $\mb{X}$ and $\mb{S}$.
We maximize a variational lower bound
 $\mathcal{L}(\theta,\beta,\gamma)$
 of the marginal log likelihood $\log p(\mb{X}|\mb{S})$.
More specifically, we have
\begin{align}
 &\log p(\mb{x}_n|\mb{s}_n)
 \nonumber\\
 &\geq
 \ \mathbb{E}_q
 [
 \log p_\theta(\mb{x}_n|\mb{s}_n,\mb{g}_n) + \log p(\mb{a}_n) + \log p(\mb{g}_n)
 \nonumber\\
 &\: \:\:\:\:\:\:\:\:\:\:
 - \log q_\beta(\mb{a}_n|\mb{s}_n,\mb{x}_n) - \log q_\gamma(\mb{g}_n|\mb{s}_n,\mb{x}_n)
 ]
 \nonumber\\
 &=
 \ \mathbb{E}_{q}
 [\log p_\theta(\mb{x}_n|\mb{s}_n,\mb{a}_n,\mb{g}_n)]
 \nonumber\\
 &\: \:\:\:\:\:\:\:\:\:\:
 - \mbox{KL}(q_\beta(\mb{a}_n|\mb{s}_n,\mb{x}_n) || p(\mb{a}_n))
 \nonumber\\
 &\: \:\:\:\:\:\:\:\:\:\:
 - \mbox{KL}(q_\gamma(\mb{g}_n|\mb{s}_n,\mb{x}_n) || p(\mb{g}_n))
 \nonumber\\
 &\overset{\mbox{\tiny def}}{=}
 \mathcal{L}_n(\theta,\beta,\gamma),
\end{align}
where $\mathcal{L}(\theta,\beta,\gamma) = \sum_{n=1}^N \mathcal{L}_n(\theta,\beta,\gamma)$.
The three networks $\theta$, $\beta$, and $\gamma$ can be optimized jointly
 by using the reparametrization tricks~\cite{kingma2014autoencoding}
 given by \eref{eq:sampling_an} and \eref{eq:sampling_gn},
 where the variance $\sigma_\theta^2(\mb{s}_n,\mb{a}_n,\mb{g}_n)$ of the generator $\theta$
  is fixed to 1 for stability.

To encourage the disentanglement
 between the foreground features (appearance) $\mb{a}_n$
 and the background features (scene) $\mb{g}_n$,
 we separately input foreground and background parts of the original image $\mb{x}_n$
 into the two encoders $\beta$ and $\gamma$, respectively,
 instead of directly feeding $\mb{x}_n$ into $\beta$ and $\gamma$.
Specifically,
 an image $\mb{x}^*_n \in \mathbb{R}^{D^{\mb{x}^*}}$, a reduced-size version of $\mb{x}_n$,
 is first split into foreground and
 background images $\mb{x}^{\mathrm{fg}}_n$ and $\mb{x}^{\mathrm{bg}}_n$ as follows:
\begin{align}
\mb{x}^{\mathrm{fg}}_n &= \mb{x}^*_n \odot \mb{w}_n, \\
\mb{x}^{\mathrm{bg}}_n &= \mb{x}^*_n \odot (\mb{1} - \mb{w}_n),
\end{align}
where $\odot$ indicates the element-wise product
 and $\mb{w}_n \in \mathbb{R}^{D^{\mb{x}^*}}$ represents a mask image
 estimated from $\mb{x}^*_n$ with the additional information of the pose $\mb{s}_n$.
In this paper,
 we use a neural mask estimator $\psi$ trained
 in a supervised manner such that the mean squared error
 between the estimated and ground-truth masks is minimized.

\newcommand{\figrecognizer}{%
  \includegraphics[width=.86\linewidth]{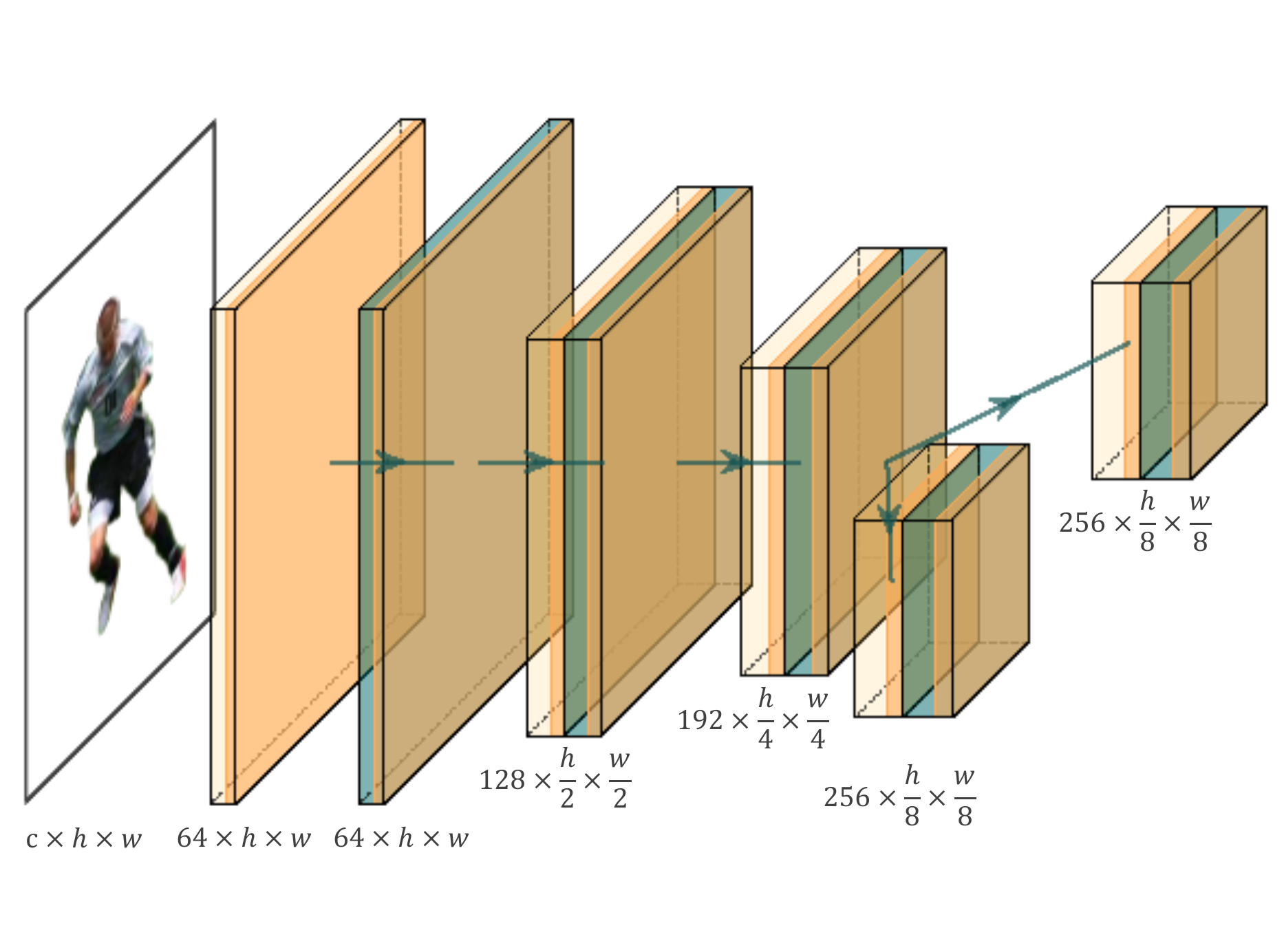}
  \caption{The network architecture common 
  in the recognizers $\beta$ and $\gamma$ of the pose-conditioned image VAE 
  taking as input foreground and background images $\mb{x}_n^{\mathrm{fg}}$
  and $\mb{x}_n^{\mathrm{bg}}$ ($c=3$), respectively,
  and the recognizer $\delta$ of the pose VAE
  taking as input the pose $\mb{s}_n$ ($c$ is the number of joints)}
  \label{fig:recognizer}
}

\newcommand{\figguide}{%
  \includegraphics[width=.86\linewidth]{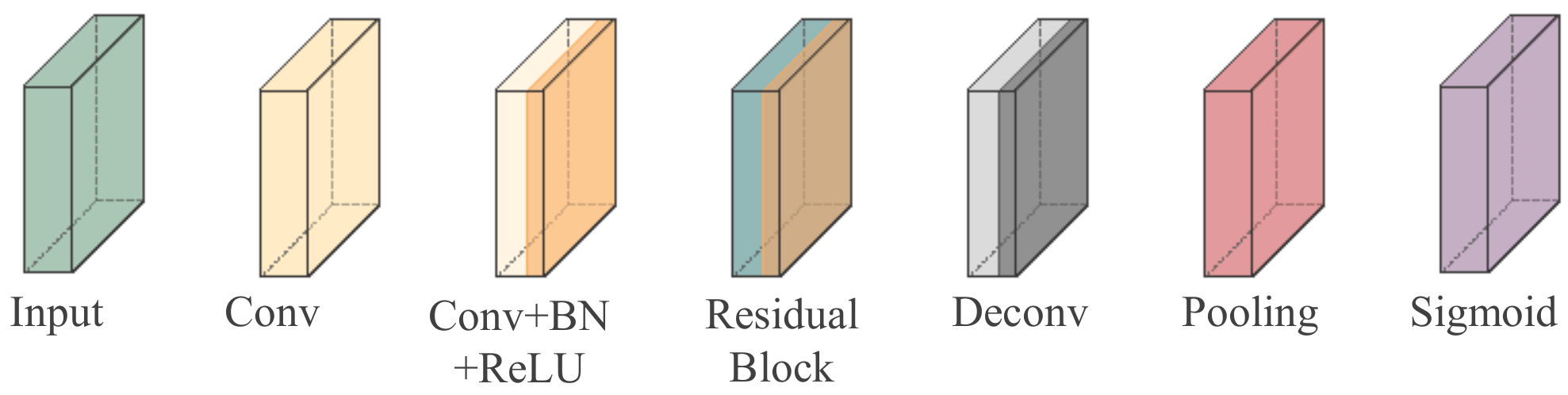}
  \caption{Layers used for implementing DNNs}
  \label{fig:guide}
}

\newcommand{\figgeneratorpose}{%
  \includegraphics[width=.7\linewidth]{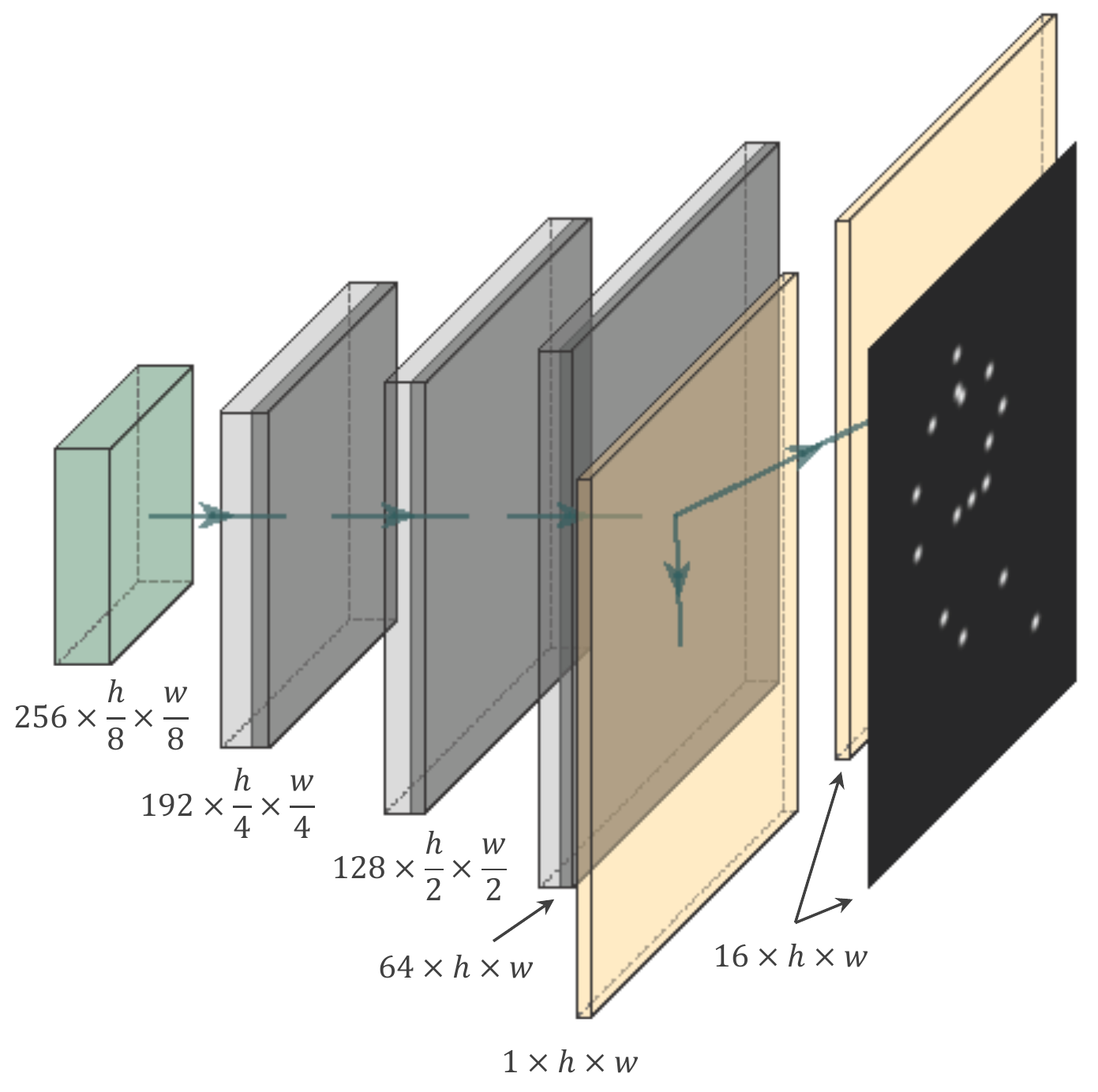}
  \caption{The network architecture of the generator $\theta$ of pose VAE
  taking as input the primitives $\mb{z}_n$
  and yielding the mean $\mu_\phi(\mb{z}_n)$ and variance $\sigma_\phi(\mb{z}_n)$}
  \label{fig:generatorpose}
}

\newcommand{\figgeneratorimage}{%
  \includegraphics[width=.84\linewidth]{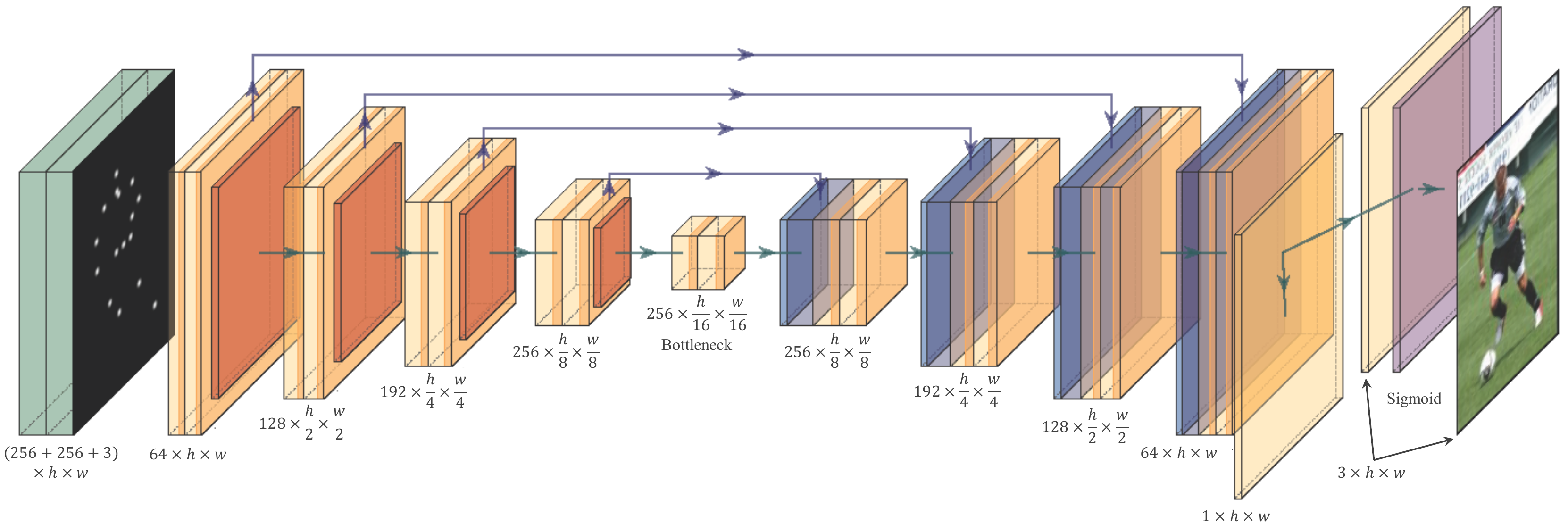}
  \caption{The network architecture of the generator $\theta$ of the pose-conditioned Image VAE
  taking as input the pose $\mb{s}_n$, the appearance $\mb{a}_n$, and the scene $\mb{g}_n$
  and yielding the mean $\mu_\theta(\mb{a}_n,\mb{g}_n,\mb{s}_n)$ and the variance $\sigma_\theta(\mb{a}_n,\mb{g}_n,\mb{s}_n)$}
  \label{fig:generatorimage}
}

\begin{figure}[t]
  \centering
  \figrecognizer
\end{figure}

\begin{figure}[t]
  \centering
  \figguide
\end{figure}

\begin{figure}[t]
  \centering
  \figgeneratorpose
\end{figure}

\begin{figure*}[t]
 \centering
 \figgeneratorimage
\end{figure*}

The recognizers $\beta$ and $\gamma$ are implemented
 as stacks of four residual blocks~\cite{he2016deep}
 (Fig.~\ref{fig:recognizer} and Fig.~\ref{fig:guide}).
Unlike the original ResNet,
 a branching architecture is introduced in the last layer
 to output the mean and variance of the posterior Gaussian distribution.
The generator $\theta$ is implemented with a U-Net~\cite{ronneberger2015u}
 that takes as input a stack of the heatmaps of the joints given by $\mb{s}_n$
 and the latent variables $\mb{a}_n$ and $\mb{g}_n$,
 where a branching architecture is introduced in the last layer
 to evaluate the pose fidelity with $\mb{x}_n$ (Fig.~\ref{fig:generatorimage}).
The mask estimator $\psi$ is implemented as a U-Net~\cite{ronneberger2015u}
 that takes as input a shrunk image $\mb{x}^*_n$
 and a stack of the heatmaps of the joints given by $\mb{s}_n$
 and outputs a mask image $\mb{w}_n$.
To obtain sharper mask images,
 we apply a sigmoid function, $\varsigma(x)=(1+\exp{(-10x)})^{-1}$, 
 to every element of the output $\mb{w}_n$ of the pre-trained estimator $\psi$. 

\subsection{Pose VAE}
\label{sec:pose_vae}

The pose VAE consisting of
 the pose-to-primitive recognizer $\delta$ and the primitive-to-pose generator $\phi$ (decoder)
 is pre-trained  in an \textit{unsupervised} manner
 by using only $\mb{S}$.
We maximize a variational lower bound
 $\mathcal{L}(\phi,\delta)$ of the marginal log likelihood $\log p(\mb{S})$
 evaluating the pose plausibility.
More specifically, we have
\begin{align}
 &\log p(\mb{s}_n)
 \nonumber\\
 &\ge
 \ \mathbb{E}_q\!
 \left[
 \log p_\phi(\mb{s}_n|\mb{z}_n) + \log p(\mb{z}_n) - \log q_\delta(\mb{z}_n|\mb{s}_n)
 \right]
 \nonumber\\
 &=
 \ \mathbb{E}_{q}
 [\log p_\phi(\mb{s}_n|\mb{z}_n)]
 -
 \mbox{KL}(q_\delta(\mb{z}_n|\mb{s}_n) || p(\mb{z}_n))
 \nonumber\\
 &\overset{\mbox{\tiny def}}{=}
 \ \mathcal{L}_n(\phi,\delta),
 \label{eq:derivation_lower_bound_pose}
\end{align}
where $\mathcal{L}(\phi,\delta) = \sum_{n=1}^N \mathcal{L}_n(\phi,\delta)$.
The two networks $\phi$ and $\delta$ can be optimized jointly
 by using the reparametrization trick~\cite{kingma2014autoencoding}
 given by \eref{eq:sampling_zn},
 where the variance $\sigma_\phi^2$ of the generator $\phi$
  is fixed to 1 for stability.

The recognizer $\delta$ is implemented
 in the same way as the recognizers $\beta$ and $\gamma$
 except that it has a different input dimension (Fig. \ref{fig:recognizer}).
The generator $\phi$ is implemented as a three-layered transposed convolutional network,
 where a branching architecture was introduced in the last layer
 to evaluate the pose plausibility (Fig. \ref{fig:generatorpose}).


\section{Evaluation}
\label{sec:evaluation}

\begin{figure}[!tb]
\centering
\includegraphics[width=0.65\linewidth]{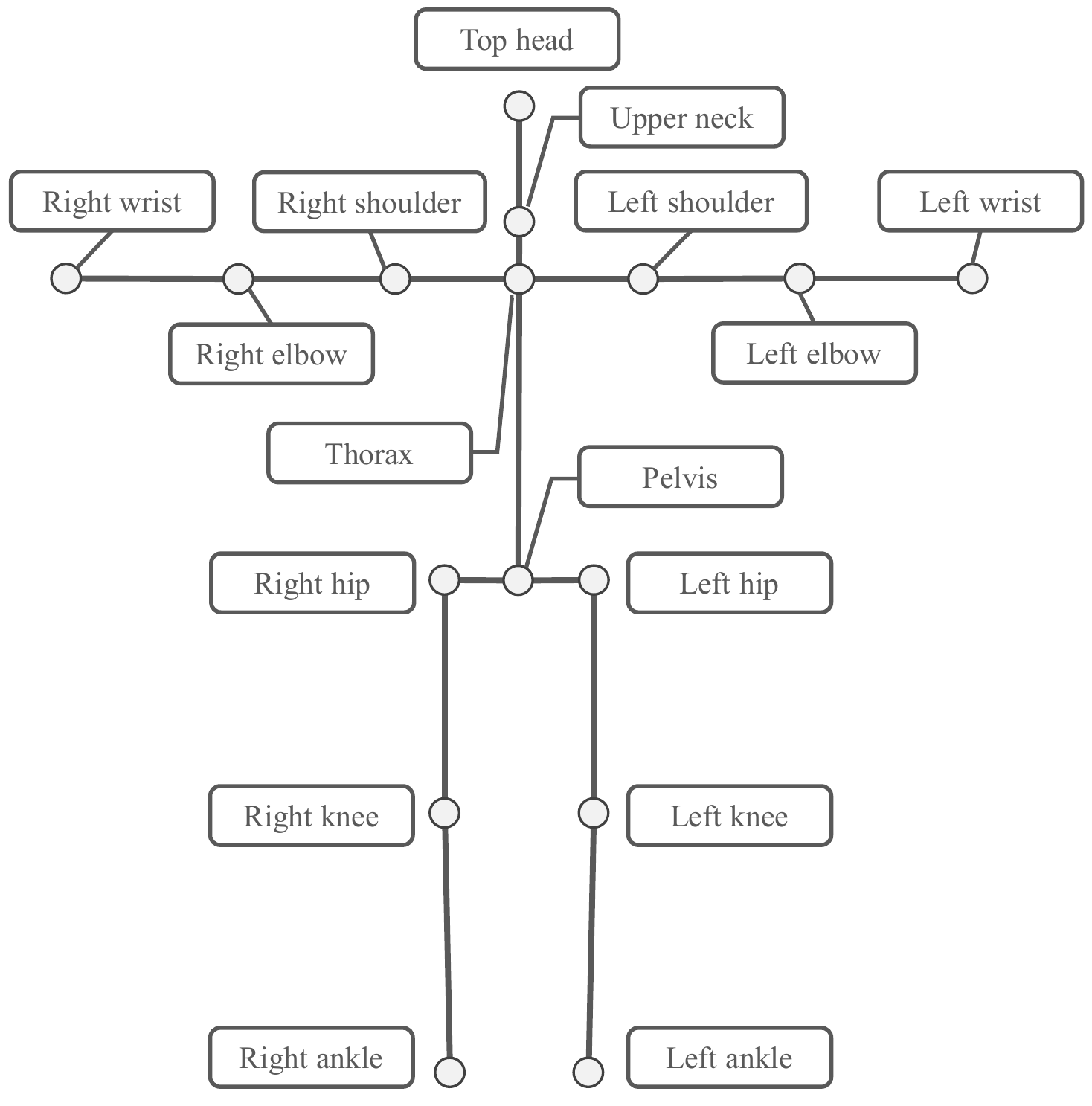}
\caption{sixteen joints considered in our experiments}
\label{fig:stick}
\end{figure}

This section reports comparative experiments
 conducted for evaluating the effectiveness
 of our semi-supervised plausibility- and fidelity-aware pose estimation method.
Our goal is to train a neural pose estimator
 that detects the coordinates of 16 joints
 (namely, right ankle, right knee, right hip, left hip, left knee, left ankle, pelvis, thorax, upper neck, head top, right wrist, right elbow, right shoulder, left shoulder, left elbow, and left wrist, as shown in Fig.~\ref{fig:stick}) from an image.
We here validate two hypotheses:
 (A)
 under a \textit{supervised} condition,
 the proposed method based on the joint training of the generative and recognition models
 outperforms conventional methods based on an image-to-pose recognition model,
 and
 (B)
 under a \textit{semi-supervised} condition,
 non-annotated images can be used for improving the performance
 thanks to the power of the mirror architecture.

\subsection{Datasets and Criteria}

We used two standard datasets
 that have widely been used in conventional studies on pose estimation.

\subsubsection{Leeds Sports Pose (LSP) Dataset}

The LSP dataset with its extension~\cite{johnson2010clustered,johnson2011learning}
 contains 12K images of sports activities (11K for training and 1K for testing) in total.
Each image originally has an annotation about the coordinates of the 14 joints
 except for the pelvis and thorax.
We estimated the coordinate of the pelvis
 by averaging the coordinates of the left and right hips
 and the coordinate of the thorax
 by averaging the coordinates of the left and right shoulders.
Each image was cropped to a square region centering on a person
 and then scaled to $D^\mb{x} = 256 \times 256$.

The performance of pose estimation was measured
 with the \textit{percentage of correct keypoints} (PCK)~\cite{yang2013articulated}.
The estimated coordinate of a joint was judged as correct
 if it was within $\tau \times \max(h, w)$ pixels around the ground-truth coordinate,
 where $\tau$ is a normalized distance, 
 and $h$ and $w$ are the height and width
 of the tightly cropped bounding box of the person, respectively.
We used $\tau = 0.2$ in our experiment.

\subsubsection{MPII Human Pose (MPII) Dataset}

The MPII dataset~\cite{andriluka14cvpr} contains around 25K images of daily activities
 (22K for training and 3K for testing).
Each image has an annotation about the coordinates of the 16 joints
 and was cropped to a square region centering on a person,
 and then scaled to $D^\mb{x} = 256 \times 192$.

The performance of pose estimation was measured
 with the \textit{percentage of correct keypoints in relation to head segment length} (PCKh)~\cite{andriluka14cvpr}.
The estimated coordinate of a joint was judged as correct
 if it was within $\tau l$ pixels around the ground-truth coordinate,
 where $\tau$ is a constant threshold, and $l$ is the head size
 corresponding to 60\% of the diagonal length of the ground-truth head bounding box.
We used $\tau = 0.5$ in our experiment.

\subsection{Training Procedures}

We regarded randomly selected 20\%, 40\%, 60\%, or 80\% of the training data as annotated images
 and the remaining part as non-annotated images.
Only the annotated images were used for supervised training
 and the whole training data were used for semi-supervised training.
As in the official implementation of \cite{sun2019deep},
 the training data were augmented with random scaling, rotation,
 and horizontal flipping~\cite{hrnet}.
The target data of $\mb{s}_n$ were made
 by stacking 16 reduced-size \textit{one-hot} images indicating the coordinates of the 16 joints
 ($D^\mb{s} = 16 \times 64 \times 64 \ \mbox{or} \ 16 \times 64 \times 48$).
In the test phase,
 a coordinate taking the maximum value
 in each of the 16 \textit{greyscale} images (heatmaps) of $\mb{s}_n$ was detected.

\begin{figure}[!tb]
\centering
\includegraphics[width=0.98\linewidth]{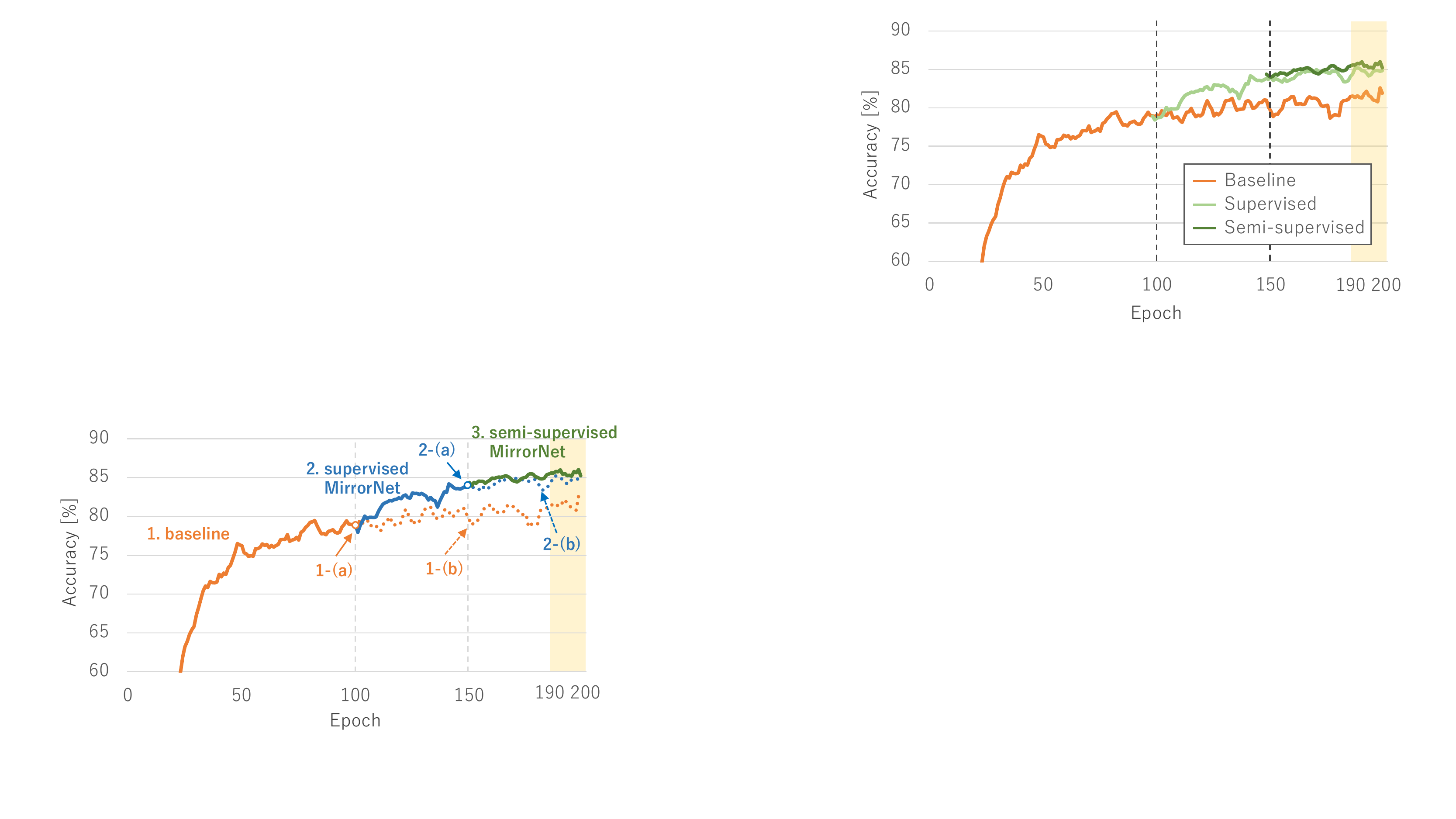}
\caption{
Learning curves with the pose recognizer $\alpha$ based on hourglass network~\cite{newell2016stacked}
on the LSP dataset (60\% of the training data were regarded as annotated images)
}
\label{fig:curve}
\end{figure}

We conducted curriculum learning as described in Section~\ref{sec:implementation} and shown in \fref{fig:curve},
 where the dimensions of the latent foreground, background and pose features
 were set to $D^\mb{a} = D^\mb{g} = D^\mb{z} = 256 \times 8 \times 8 \ \mbox{or} \ 256 \times 8 \times 6$ (\fref{fig:curve}).

\begin{enumerate}
\item
The six sub-networks were trained independently in a \textit{supervised} manner
 with the annotated images.
The pose recognizer $\alpha$ based on
 the residual hourglass network~\cite{newell2016stacked},
 ResNet-50~\cite{xiao2018simple},
 or high-resolution subnetworks~\cite{sun2019deep}
 was trained for 100 epochs
 (\sref{sec:pose_recognizer}).
The pose-conditioned image VAE
 consisting of the generator $\theta$ and the recognizers $\beta$ and $\gamma$
 was trained for 200 epochs (\sref{sec:image_vae}).
The pose VAE consisting of the generator $\phi$ and the recognizer $\delta$
 was also trained for 200 epochs (\sref{sec:pose_vae}).
The mask estimator $\psi$ was trained
 by using the UPi-S1h dataset~\cite{Lassner:up2017}
 containing human images with silhouette annotations (\ie, mask images),
 where images included in the LSP or MPII datasets were excluded.
\begin{enumerate}
\item
MirrorNet was initialized
 by combining the six sub-networks and the mask estimator $\psi$ for the step 2.
\item
The pose recognizer $\alpha$ was further trained for 100 epochs (\ie, 200 epochs in total)
 and the parameters obtained at last 10 epochs were used for evaluation
 (\textbf{baseline}).
\end{enumerate}

\item
MirrorNet was trained in a \textit{supervised} manner
 with the same annotated images for 50 epochs,
 where the mask estimator $\psi$ was not updated.
\begin{enumerate}
\item
MirrorNet obtained at the last epoch was preserved for the step 3.
\item
MirrorNet was further trained for 50 epochs
 and the parameters of the pose recognizer $\alpha$
 obtained at last 10 epochs were used for evaluation
 (\textbf{supervised MirrorNet}).
\end{enumerate}

\item
MirrorNet was further trained in a \textit{semi-supervised} manner
 with the annotated and non-annotated images for 50 epochs,
 where the mask estimator $\psi$ was not updated.
The parameters obtained at last 10 epochs were used for evaluation
 (\textbf{semi-supervised MirrorNet}).
\end{enumerate}
For a fair comparison,
 the pose recognizer $\alpha$ was trained for 200 epochs
 in total in each of the three methods.
The performance of pose estimation
 was measured by averaging PCK@0.2 or PCKh@0.5 over the last 10 epochs.

All networks were implemented using PyTorch~\cite{paszke2019pytorch}
 and optimized using Adam \cite{kingma2014adam} with a learning rate of 1e-3.
The mini-batch size was always set to 128 images,
 which had annotations in the supervised training phase
 or consisted of 96 annotated images and 32 non-annotated images in the semi-supervised training phase.
We used AI Bridging Cloud Infrastructure (ABCI)
 of National Institute of Advanced Industrial Science and Technology (AIST)
 for computation (\tref{tab:cresource}).

\begin{table}[tb]
  \centering
  \caption{Computational resource}
  \vspace{2pt}
  \begin{tabular}{@{}llc@{}}
    \toprule
    Item                 & Description                      & \#                 \\
    \midrule
    \multirow{2}{*}{CPU} & Intel Xeon Gold 6148 Processor   & \multirow{2}{*}{2} \\
                         & (2.4 GHz, 20 Cores, 40 Threads)  &                    \\
    GPU                  & NVIDIA Tesla V100 for NVLink     & 4                  \\
    Memory               & 384 GiB DDR4 2666 MHz RDIMM      &                    \\
    SSD                  & Intel SSD DC P4600 1.6 TB u.2    & 1                  \\
    Interconnects        & InfiniBand EDR (12.5 GB/s)       & 2                  \\
    \bottomrule
  \end{tabular}
  \label{tab:cresource}
\end{table}

\subsection{Experimental Results}
\label{sec:experiment_1}

\begin{table*}[htp]
    \centering
    \def\tabbullet{\hspace{1em}\textbullet\hspace{0.5em}}
    \renewcommand{\arraystretch}{1.15}
    \tabcolsep = 1.5mm
    \caption{Pose estimation performances on the LSP dataset \cite{johnson2010clustered,johnson2011learning} with the recognition model by Newell \etal\ \cite{newell2016stacked}}
    \begin{tabular}{@{}lcccccccccc@{}}
    \toprule
    & \multicolumn{2}{c}{Training data}    & \multicolumn{8}{c}{ PCK@0.2} \\
    \cmidrule(r){2-3} \cmidrule(l){4-11}
                    & \#annotated & \#non-annotated & Head  & Shoulder & Elbow & Wrist & Hip   & Knee  & Ankle & Total \\
    \midrule
    Baseline \cite{newell2016stacked} 
    & 2200 & \notapplicable & 92.24 & 80.08 & 72.57 & 69.37 & 68.36 & 68.83 & 65.95 & 74.12 \\
    Supervised MirrorNet
    & 2200 & \notapplicable & 92.94 & 83.01 & 75.30 & 72.55 & 71.81 & 72.50 & 69.40 & \textbf{76.99} \\
    Semi-supervised MirrorNet 
    & 2200 & 8800 & 89.33 & 80.70 & 71.74 & 69.64 & 65.75 & 69.24 & 67.11 & 73.64 \\
    \midrule
    Baseline \cite{newell2016stacked}
    & 4400 & \notapplicable &  94.05 & 85.09 & 77.70 & 74.79 & 75.97 & 74.69 & 71.18 & 79.27 \\
    Supervised MirrorNet 
    & 4400 & \notapplicable & 94.50 & 87.31 & 81.71 & 79.04 & 78.71 & 78.80 & 75.44 & 82.39 \\
    Semi-supervised MirrorNet 
    & 4400 & 6600 & 93.66 & 88.12 & 82.78 & 80.44 & 78.24 & 80.30 & 77.23 & \textbf{83.15} \\
    \midrule
    Baseline \cite{newell2016stacked} 
    & 6600 & \notapplicable &  94.43 & 86.00 & 80.13 & 77.63 & 77.55 & 78.14 & 74.02 & 81.31 \\
    Supervised MirrorNet 
    & 6600 & \notapplicable & 94.87 & 88.37 & 83.57 & 81.19 & 80.24 & 82.77 & 79.23 & 84.49 \\
    Semi-supervised MirrorNet 
    & 6600 & 4400 & 94.97 & 88.74 & 84.42 & 82.51 & 80.78 & 84.08 & 81.10 & \textbf{85.39} \\
    \midrule
    Baseline \cite{newell2016stacked}
    & 8800 & \notapplicable & 94.63 & 86.78 & 80.56 & 78.91 & 77.46 & 79.58 & 75.46 & 82.11 \\
    Supervised MirrorNet
    & 8800 & \notapplicable & 95.21 & 89.91 & 85.60 & 83.85 & 81.80 & 84.35 & 82.69 & \textbf{86.38} \\
    Semi-supervised MirrorNet
    & 8800 & 2200 & 95.34 & 89.04 & 84.37 & 83.58 & 81.23 & 85.07 & 83.05 & 86.15 \\
    \bottomrule
    \end{tabular}
    \label{tab:result_nl}
\end{table*}

\begin{table*}[htp]
    \centering
    \def\tabbullet{\hspace{1em}\textbullet\hspace{0.5em}}
    \renewcommand{\arraystretch}{1.15}
    \tabcolsep = 1.5mm
    \caption{Pose estimation performances on the LSP dataset \cite{johnson2010clustered,johnson2011learning} with the recognition model by Xiao \etal\ \cite{xiao2018simple}}
    \begin{tabular}{@{}lcccccccccc@{}}
    \toprule
    & \multicolumn{2}{c}{Training data}    & \multicolumn{8}{c}{PCK@0.2} \\
    \cmidrule(r){2-3} \cmidrule(l){4-11}
                    & \#annotated & \#non-annotated & Head  & Shoulder & Elbow & Wrist & Hip   & Knee  & Ankle & Total \\
    \midrule
    Baseline \cite{xiao2018simple}
    & 2200 & \notapplicable & 85.33 & 67.68 & 54.39 & 51.96 & 55.06 & 51.68 & 47.66 & 59.48 \\
    Supervised MirrorNet 
    & 2200 & \notapplicable & 88.96 & 75.92 & 66.18 & 61.43 & 62.22 & 61.42 & 53.83 & \textbf{67.44} \\
    Semi-supervised MirrorNet 
    & 2200 & 8800 & 86.46 & 75.54 & 66.09 & 61.80 & 60.92 & 62.00 & 51.95 & 66.73 \\
    \midrule
    Baseline \cite{xiao2018simple}
    & 4400 & \notapplicable & 87.63 & 74.99 & 65.36 & 60.09 & 63.72 & 61.99 & 52.66 & 66.94 \\
    Supervised MirrorNet 
    & 4400 & \notapplicable & 89.94 & 80.47 & 72.19 & 67.87 & 69.50 & 69.41 & 60.20 & \textbf{73.06} \\
    Semi-supervised MirrorNet 
    & 4400 & 6600 & 88.83 & 79.62 & 71.33 & 67.43 & 67.54 & 67.78 & 60.09 & 72.07 \\
    \midrule
    Baseline \cite{xiao2018simple}
    & 6600 & \notapplicable & 90.02 & 79.03 & 69.58 & 64.22 & 69.28 & 64.99 & 56.04 & 70.69 \\
    Supervised MirrorNet 
    & 6600 & \notapplicable & 91.88 & 82.65 & 75.02 & 71.89 & 73.39 & 73.27 & 65.51 & \textbf{76.49} \\
    Semi-supervised MirrorNet 
    & 6600 & 4400 & 91.68 & 82.80 & 75.63 & 71.99 & 73.29 & 71.87 & 65.31 & 76.31 \\
    \midrule
    Baseline \cite{xiao2018simple} 
    & 8800 & \notapplicable & 87.78 & 76.66 & 66.30 & 60.24 & 66.75 & 61.85 & 55.23 & 68.17 \\
    Supervised MirrorNet 
    & 8800 & \notapplicable & 91.91 & 83.34 & 75.76 & 71.48 & 72.57 & 72.24 & 64.97 & \textbf{76.31} \\
    Semi-supervised MirrorNet 
    & 8800 & 2200 & 91.28 & 82.14 & 74.71 & 70.83 & 71.66 & 70.89 & 63.38 & 75.26 \\
    \bottomrule
    \end{tabular}
    \label{tab:result_xl}
\end{table*}

\begin{table*}[htp]
    \centering
    \def\tabbullet{\hspace{1em}\textbullet\hspace{0.5em}}
    \renewcommand{\arraystretch}{1.15}
    \tabcolsep = 1.5mm
    \caption{Pose estimation performances on the LSP dataset \cite{johnson2010clustered,johnson2011learning} with the recognition model by Sun \etal\ \cite{sun2019deep}}
    \begin{tabular}{@{}lcccccccccc@{}}
    \toprule
    & \multicolumn{2}{c}{Training data}    & \multicolumn{8}{c}{PCK@0.2} \\
    \cmidrule(r){2-3} \cmidrule(l){4-11}
                    & \#annotated & \#non-annotated & Head  & Shoulder & Elbow & Wrist & Hip   & Knee  & Ankle & Total \\
    \midrule
    Baseline \cite{sun2019deep}
    & 2200 & \notapplicable & 92.15 & 81.26 & 72.92 & 71.26 & 69.69 & 70.32 & 67.85 & 75.27 \\
    Supervised MirrorNet 
    & 2200 & \notapplicable & 93.45 & 84.00 & 77.27 & 75.78 & 73.96 & 74.23 & 72.28 & \textbf{78.91} \\
    Semi-supervised MirrorNet
    & 2200 & 8800 & 90.28 & 82.99 & 75.75 & 73.94 & 67.85 & 75.07 & 72.79 & 77.25 \\
    \midrule
    Baseline \cite{sun2019deep}
    & 4400 & \notapplicable & 93.17 & 84.51 & 77.24 & 74.62 & 74.54 & 74.63 & 72.70 & 79.00 \\
    Supervised MirrorNet & 4400 & \notapplicable & 95.04 & 87.80 & 81.82 & 79.59 & 79.30 & 81.56 & 79.19 & 83.65 \\
    Semi-supervised MirrorNet & 4400 & 6600 & 93.94 & 87.72 & 82.67 & 80.57 & 77.80 & 81.65 & 79.92 & \textbf{83.70} \\
    \midrule
    Baseline \cite{sun2019deep} 
    & 6600 & \notapplicable & 93.06 & 84.76 & 77.50 & 74.46 & 74.11 & 76.36 & 73.47 & 79.36 \\
    Supervised MirrorNet 
    & 6600 & \notapplicable & 95.01 & 88.94 & 83.63 & 81.67 & 80.49 & 83.53 & 80.75 & 85.06 \\
    Semi-supervised MirrorNet 
    & 6600 & 4400 & 95.11 & 88.67 & 83.68 & 82.51 & 81.00  & 84.45 & 81.83 & \textbf{85.51} \\
    \midrule
    Baseline \cite{sun2019deep} 
    & 8800 & \notapplicable & 94.05 & 85.22 & 78.16 & 76.06 & 75.17 & 77.69 & 75.66 & 80.51 \\
    Supervised MirrorNet 
    & 8800 & \notapplicable & 95.13 & 88.56 & 84.09 & 83.21 & 80.76 & 85.10 & 83.30 & 85.96 \\
    Semi-supervised MirrorNet 
    & 8800 & 2200 & 95.34 & 89.04 & 84.37 & 83.58 & 81.23 & 85.07 & 83.05 & \textbf{86.15} \\
    \bottomrule
    \end{tabular}
    \label{tab:result_sl}
\end{table*}

\begin{table*}[htp]
    \centering
    \renewcommand{\arraystretch}{1.15}
    \tabcolsep = 1.5mm
    \caption{Pose estimation performances on the MPII dataset \cite{andriluka14cvpr} with the recognition model by Newell \etal\ \cite{newell2016stacked}}
    \begin{tabular}{@{}lcccccccccc@{}}
    \toprule
    & \multicolumn{2}{c}{Training data}    & \multicolumn{8}{c}{PCKh@0.5} \\
    \cmidrule(r){2-3} \cmidrule(l){4-11}
                    & \#annotated & \#non-annotated & Head  & Shoulder & Elbow & Wrist & Hip   & Knee  & Ankle & Total \\
    \midrule
    Baseline \cite{newell2016stacked} 
    & 4449 & \notapplicable & 93.26 & 86.86 & 76.10 & 69.17 & 75.07 & 67.82 & 63.91 & 76.91 \\
    Supervised MirrorNet & 4449 & \notapplicable & 94.10 & 89.27 & 79.05 & 72.89 & 77.64 & 71.16 & 67.60 & \textbf{79.63} \\
     Semi-supervised MirrorNet & 4449 & 17797 & 93.34 & 87.92 & 77.00 & 70.03 & 73.60 & 67.68 & 63.02 & 77.02 \\
    \midrule
    Baseline \cite{newell2016stacked} 
    & 8899 & \notapplicable & 94.32 & 89.07 & 78.58 & 71.36 & 78.48 & 71.29 & 67.29 & 79.43 \\
     Supervised MirrorNet & 8899 & \notapplicable & 95.12 & 91.40 & 81.97 & 75.57 & 81.98 & 75.14 & 71.20 & 82.51 \\
     Semi-supervised MirrorNet & 8899 & 13347 & 95.19 & 92.15 & 83.27 & 77.03 & 81.57 & 76.30 & 72.82 & \textbf{83.32} \\
    \midrule
    Baseline \cite{newell2016stacked} 
    & 13347 & \notapplicable & 94.93 & 91.24 & 81.63 & 74.97 & 81.92 & 74.41 & 69.83 & 82.07 \\
     Supervised MirrorNet & 13347 & \notapplicable & 95.92 & 93.37 & 84.96 & 78.54 & 85.07 & 78.80 & 74.90 & 85.18 \\
     Semi-supervised MirrorNet & 13347 & 8899 & 95.85 & 93.56 & 85.06 & 79.15 & 85.35 & 79.26 & 75.47 & \textbf{85.47} \\
    \midrule
    Baseline \cite{newell2016stacked}
    & 17797 & \notapplicable & 94.90 & 91.32 & 81.62 & 74.70 & 81.06 & 74.39 & 70.32 & 81.94 \\
     Supervised MirrorNet & 17797 & \notapplicable & 95.85 & 93.45 & 85.40 & 79.32 & 85.24 & 79.32 & 75.69 & \textbf{85.54} \\
     Semi-supervised MirrorNet & 17797 & 4449 & 95.76 & 93.36 & 85.14 & 79.21 & 84.79 & 79.29 & 75.64 & 85.38 \\
    \bottomrule
    \end{tabular}
    \label{tab:result_nm}
\end{table*}

\begin{table*}[htp]
    \centering
    \renewcommand{\arraystretch}{1.15}
    \tabcolsep = 1.5mm
    \caption{Pose estimation performances on the MPII dataset \cite{andriluka14cvpr} with the recognition model by Xiao \etal\ \cite{xiao2018simple}}
    \begin{tabular}{@{}lcccccccccc@{}}
    \toprule
    & \multicolumn{2}{c}{Training data}    & \multicolumn{8}{c}{PCKh@0.5} \\
    \cmidrule(r){2-3} \cmidrule(l){4-11}
                    & \#annotated & \#non-annotated & Head  & Shoulder & Elbow & Wrist & Hip   & Knee  & Ankle & Total \\
    \midrule
    Baseline \cite{xiao2018simple}
    & 4449 & \notapplicable & 88.71 & 80.11 & 64.74 & 54.85 & 66.30 & 54.39 & 52.14 & 67.10 \\
     Supervised MirrorNet & 4449 & \notapplicable & 90.71 & 82.99 & 69.26 & 60.25 & 70.08 & 59.96 & 56.31 & \textbf{71.03} \\
     Semi-supervised MirrorNet & 4449 & 17797 & 89.74 & 82.29 & 67.98 & 58.50 & 66.54 & 57.14 & 53.70 & 69.16 \\
    \midrule
    Baseline \cite{xiao2018simple} 
    & 8899 & \notapplicable & 90.56 & 82.91 & 68.39 & 58.58 & 70.61 & 59.16 & 55.31 & 70.50 \\
     Supervised MirrorNet & 8899 & \notapplicable & 92.68 & 87.16 & 74.86 & 65.74 & 76.39 & 66.54 & 61.74 & \textbf{76.01} \\
     Semi-supervised MirrorNet & 8899 & 13347 & 92.94 & 87.31 & 74.76 & 66.50 & 75.14 & 66.18 & 61.26 & 75.87 \\
    \midrule
    Baseline \cite{xiao2018simple} 
    & 13347 & \notapplicable & 90.64 & 83.58 & 68.52 & 58.19 & 71.63 & 59.38 & 55.62 & 70.79 \\
     Supervised MirrorNet & 13347 & \notapplicable & 92.96 & 87.84 & 74.80 & 65.40 & 77.32 & 66.72 & 62.22 & 76.32 \\
     Semi-supervised MirrorNet & 13347 & 8899 & 93.02 & 88.14 & 75.28 & 66.15 & 77.89 & 67.53 & 62.67 & \textbf{76.79} \\
    \midrule
    Baseline \cite{xiao2018simple} 
    & 17797 & \notapplicable & 89.49 & 81.37 & 65.70 & 54.90 & 68.78 & 56.91 & 53.50 & 68.40 \\
     Supervised MirrorNet & 17797 & \notapplicable & 92.36 & 86.68 & 73.50 & 64.19 & 76.36 & 65.63 & 60.34 & 75.19 \\
     Semi-supervised MirrorNet & 17797 & 4449 & 92.72 & 87.58 & 74.04 & 64.20 & 76.88 & 65.92 & 60.65 & \textbf{75.61} \\
    \bottomrule
    \end{tabular}
    \label{tab:result_xm}
\end{table*}

\begin{table*}[htp]
    \centering
    \renewcommand{\arraystretch}{1.15}
    \tabcolsep = 1.5mm
    \caption{Pose estimation performances on the MPII dataset \cite{andriluka14cvpr} with the recognition model by Sun \etal\ \cite{sun2019deep}}
    \begin{tabular}{@{}lcccccccccc@{}}
    \toprule
    & \multicolumn{2}{c}{Training data}    & \multicolumn{8}{c}{PCKh@0.5} \\
    \cmidrule(r){2-3} \cmidrule(l){4-11}
                    & \#annotated & \#non-annotated & Head  & Shoulder & Elbow & Wrist & Hip   & Knee  & Ankle & Total \\
    \midrule
    Baseline \cite{sun2019deep} 
    & 4449 & \notapplicable & 92.89 & 86.95 & 75.44 & 68.99 & 74.50 & 66.80 & 62.73 & 76.42 \\
     Supervised MirrorNet & 4449 & \notapplicable & 93.89 & 89.40 & 79.90 & 73.70 & 77.76 & 72.07 & 67.74 & \textbf{80.04} \\
     Semi-supervised MirrorNet & 4449 & 17797 & 93.92 & 88.89 & 79.05 & 72.40 & 75.14 & 70.59 & 66.22 & 78.91 \\
    \midrule
    Baseline \cite{sun2019deep} 
    & 8899 & \notapplicable & 93.86 & 88.27 & 77.52 & 70.69 & 77.62 & 69.71 & 66.05 & 78.52 \\
     Supervised MirrorNet & 8899 & \notapplicable & 95.32 & 91.82 & 83.09 & 76.67 & 82.71 & 76.44 & 72.43 & \textbf{83.35}  \\
     Semi-supervised MirrorNet & 8899 & 13347 & 95.32 & 91.97 & 83.09 & 76.77 & 81.31 & 76.08 & 72.26 & 83.12 \\
    \midrule
    Baseline \cite{sun2019deep} 
    & 13347 & \notapplicable & 93.83 & 88.23 & 78.02 & 70.78 & 77.87 & 69.92 & 66.26 & 78.70 \\
     Supervised MirrorNet & 13347 & \notapplicable & 95.84 & 92.67 & 84.60 & 78.04 & 84.03 & 77.62 & 73.76 & 84.48 \\
     Semi-supervised MirrorNet & 13347 & 8899 & 95.67 & 92.59 & 84.67 & 78.21 & 84.15 & 77.95 & 73.90 & \textbf{84.57} \\
    \midrule
    Baseline \cite{sun2019deep} 
    & 17797 & \notapplicable & 93.51 & 87.55 & 76.84 & 69.50 & 75.97 & 67.84 & 63.48 & 77.31 \\
     Supervised MirrorNet & 17797 & \notapplicable & 95.71 & 92.93 & 84.31 & 77.73 & 83.89 & 77.41 & 73.16 & \textbf{84.30} \\
     Semi-supervised MirrorNet & 17797 & 4449 & 95.60 & 92.81 & 84.31 & 77.55 & 84.09 & 77.50 & 73.26 & \textbf{84.30} \\
    \bottomrule
    \end{tabular}
    \label{tab:result_sm}
\end{table*}

\begin{figure*}[!tb]
\centering
\includegraphics[width=0.98\linewidth]{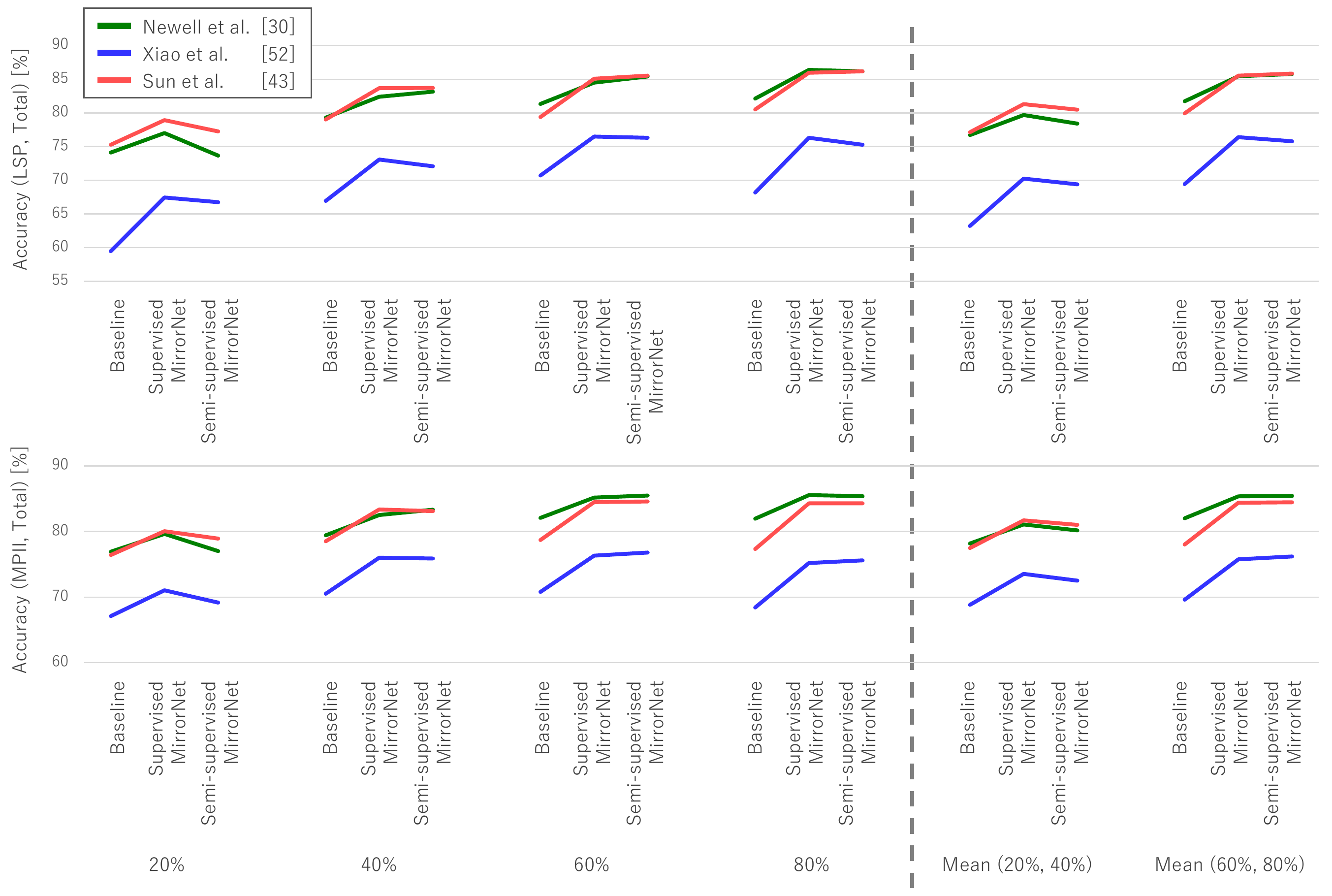}
\caption{
Pose estimation performances 
 (``Total'' columns of Tables \ref{tab:result_nl}--\ref{tab:result_sm}).
The top and bottom rows show the performances 
 on the LSP dataset~\cite{johnson2010clustered,johnson2011learning} 
 and the MPII dataset~\cite{andriluka14cvpr}, respectively.
From left to right,
 the first four columns show the performances under the conditions that 
 20\%, 40\%, 60\%, and 80\% of the training data were regarded as annotated images, respectively, 
 and the last two columns show the average performances under the 20\% and 40\% conditions
 and those under the 60\% and 80\% conditions, respectively
}
\label{fig:experiment1}
\end{figure*}

Tables \ref{tab:result_nl}--\ref{tab:result_sl}
and Tables \ref{tab:result_nm}--\ref{tab:result_sm}
 show the performances of pose estimation
 obtained by the pose recognizer $\alpha$ (\cite{newell2016stacked}, \cite{xiao2018simple}, or \cite{sun2019deep}))
 trained in the three ways (baseline, supervised MirrorNet, and semi-supervised MirrorNet)
 on the LSP and MPII datasets, respectively,
 and \fref{fig:experiment1} comparatively show the performances
 listed in the ``Total'' columns of Tables \ref{tab:result_nl}--\ref{tab:result_sm}.
In any condition, the supervised MirrorNet outperformed the baseline method
 by $5.08\pm1.71$ points on the LSP dataset
 and $4.62\pm1.42$ points on the MPII dataset,
 where the means and standard deviations were computed over the twelve conditions,
 \ie, all possible combinations of the pose recognizers \cite{newell2016stacked,xiao2018simple,sun2019deep}
 and the four ratios of annotated images (20\%, 40\%, 60\%, and 80\%).
The left four columns of \fref{fig:experiment1}
 clearly show that the supervised MirrorNet significantly outperformed the baseline method.
This strongly supports the hypothesis (A);
 the joint training of the generative and recognition models
 leads to performance improvement.
The fidelity and plausibility of estimated poses,
 which were evaluated by the pose-to-image generator $\theta$ and the pose VAE, respectively,
 were key factors for accurate pose estimation.

\begin{figure*}[!tb]
\centering
\includegraphics[width=\linewidth]{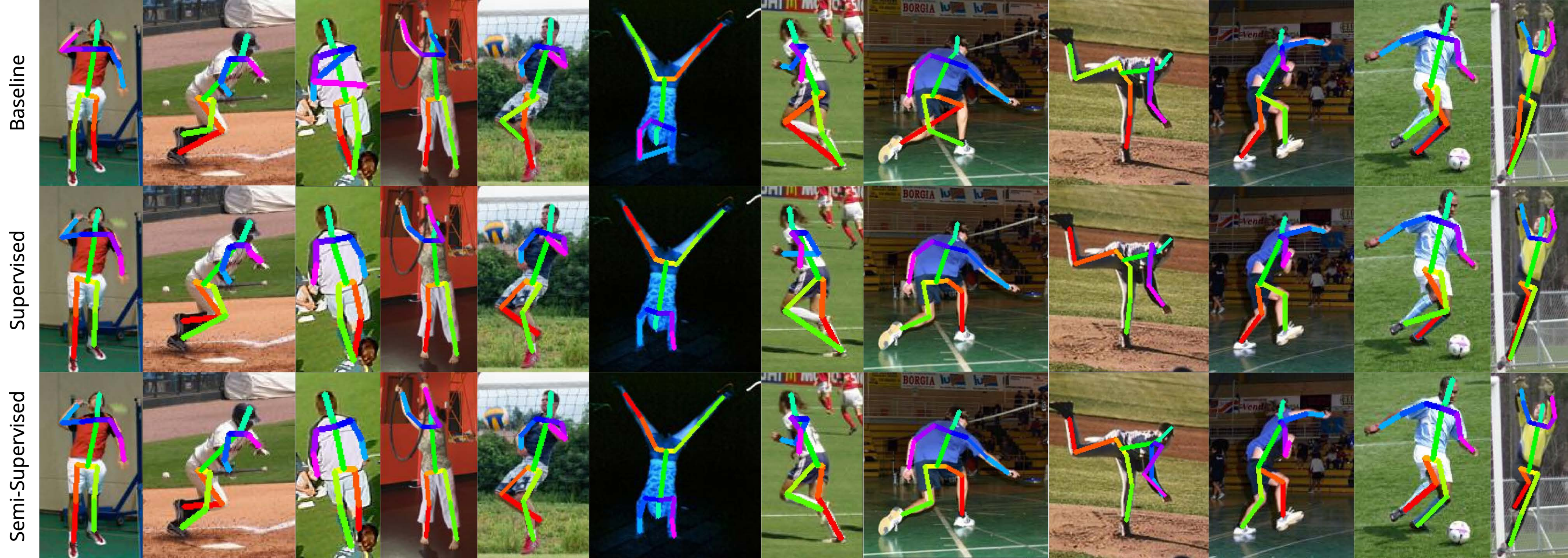}
\caption{Examples of pose estimation obtained by the baseline method~\cite{newell2016stacked},
the supervised and semi-supervised versions of MirrorNet.
Anatomically implausible poses were corrected by the MirrorNet architecture}
\label{fig:results}
\end{figure*}

We found that the semi-supervised MirrorNet outperformed the supervised MirrorNet
 when the ratios of annotated images were higher in the training data.
As shown in the right two columns of \fref{fig:experiment1},
 the semi-supervised MirrorNet tended to outperform the supervised MirrorNet
 when the annotation ratio was 60\% or 80\%.
The only exception was the condition
 that the pose recognizer $\alpha$
 was implemented with ResNet-50~\cite{xiao2018simple} on the LSP dataset.
Because the performance of this pose recognizer was insufficient,
 the pose-to-image generator $\theta$ and the pose VAE
 cannot be updated appropriately
 by using non-annotated images with estimated poses.
When the annotation ratio was 20\% or 40\%,
 the semi-supervised MirrorNet underperformed the supervised MirrorNet.
In these conditions,
 the pose-to-image generator $\theta$ and the pose VAE
 could not appropriately evaluate the fidelity and plausibility of estimated poses,
 \ie, gave wrong feedback to the pose recognizer $\alpha$ in the steps 2 and 3 of curriculum learning,
 leading to the performance degradation of the semi-supervised MirrorNet.
These results conditionally support the hypothesis (B);
 the semi-supervised training method helps
 if the performance of the conventional supervised method is above a certain level.

As shown in \fref{fig:results},
 the pose recognizer $\alpha$ trained by using the MirrorNet architecture
 yielded anatomically plausible poses.
For a better understanding of how each part of MirrorNet works,
 we show examples of person images generated by the pose-conditioned VAE in Fig.~\ref{fig:genimage},
 pose images by the pose VAE in Fig.~\ref{fig:genpose},
 and silhouette images by the mask estimator $\phi$ in Fig.~\ref{fig:genmask}
 in the appendix.
As shown in \tref{tab:ccost},
 the training of the whole MirrorNet is computationally demanding
 because the generative and recognition models of pose and images should be trained jointly.
Note that
 only the pose recognizer $\alpha$ is used in the runtime;
 the pose-conditioned VAE and the pose VAE serve as regularizers
 that stabilize the training of the MirrorNet.

\begin{table*}[!tb]
    \centering
    \def\tabbullet{\hspace{1em}\textbullet\hspace{0.5em}}
    \caption{Network size and computation speed}
    \vspace{2pt}
    \begin{tabular}{@{}l|rcc@{}}
    \toprule
    Network & \#params & GFLOPs (LSP) & GFLOPs (MPII) \\
    \midrule
    \textbf{Pose recognizer $\alpha$}       & \multicolumn{3}{c}{}   \\
     -- Hourglass~\cite{newell2016stacked}  & 25.59M & 26.17 & 19.62 \\
     -- ResNet-50~\cite{xiao2018simple}     & 34.00M & 11.99 & 8.99  \\
     -- HRNet~\cite{sun2019deep}            & 28.54M & 9.49 & 7.12  \\
    \textbf{Pose-conditioned image VAE}
                                            & \multicolumn{3}{c}{}   \\
     -- Appearance and scene recognizers $\beta$ and $\gamma$       
                                            & 5.28M  & 1.11 & 0.83  \\
     -- Image generator $\theta$                  
                                            & 12.84M & 3.25 & 2.44  \\
     -- Mask estimator $\psi$               & 10.28M & 1.42 & 1.06  \\
    \textbf{Pose VAE} & \multicolumn{3}{c}{}   \\
     -- Primitive recognizer $\delta$     & 5.29M  & 1.13 & 0.85  \\
     -- Pose generator $\phi$        & 1.33M  & 1.13 & 0.85  \\
    \midrule
    \textbf{MirrorNet (training)}       & \multicolumn{3}{c}{}   \\
     -- $\alpha$ (Hourglass~\cite{newell2016stacked}), $\beta$, $\gamma$, $\delta$, $\theta$, $\phi$, and $\psi$ & 65.89M & 35.32 & 26.48 \\
     -- $\alpha$ (ResNet-50~\cite{xiao2018simple}), $\beta$, $\gamma$, $\delta$, $\theta$, $\phi$, and $\psi$ & 74.30M & 21.14 & 15.85  \\
     -- $\alpha$ (HRNet~\cite{sun2019deep}), $\beta$, $\gamma$, $\delta$, $\theta$, $\phi$, and $\psi$           & 68.84M & 18.64 & 13.98  \\
    \textbf{MirrorNet (runtime)}        & \multicolumn{3}{c}{}   \\
     -- only $\alpha$ (Hourglass~\cite{newell2016stacked}) 
                                            & 25.59M & 26.17 & 19.62 \\
     -- only $\alpha$ (ResNet-50~\cite{xiao2018simple})   
                                            & 34.00M & 11.99 & 8.99  \\
     -- only $\alpha$ (HRNet~\cite{sun2019deep})           
                                            & 28.54M & 9.49 & 7.12  \\
    \bottomrule
    \end{tabular}
    \label{tab:ccost}
\end{table*}

\begin{table*}[htp]
    \centering
    \renewcommand{\arraystretch}{1.15}
    \tabcolsep = 1.5mm
    \caption{Pose estimation performances with respect to the ratio of annotated images in each mini-batch}
    \begin{tabular}{@{}lcccccccccc@{}}
    \toprule
    & \multicolumn{2}{c}{Mini-batch composition} & \multicolumn{8}{c}{PCK@0.2} \\
    \cmidrule(r){2-3} \cmidrule(l){4-11}
    & \#annotated & \#non-annotated & Head  & Shoulder & Elbow & Wrist & Hip & Knee  & Ankle & Total \\
    \midrule
    Supervised MirrorNet & 128 & \notapplicable & 94.02 & 83.68 & 76.36 & 75.56 & 73.17 & 74.97 & 72.12 & 78.73 \\
    \midrule
    Semi-supervised MirrorNet & 32 & 96 & 93.42 & 84.65 & 78.34 & 76.43 & 73.61 & 75.91 & 72.19 & \textbf{79.44} \\
    & 48  & 80 & 93.34 & 83.20 & 77.23 & 76.55 & 72.79 & 74.94 & 72.18 & 78.84 \\
    & 64  & 64 & 93.19 & 83.99 & 77.83 & 77.39 & 73.44 & 75.67 & 72.95 & \textbf{79.44} \\
    & 80  & 48 & 90.79 & 82.82 & 77.28 & 75.71 & 71.53 & 74.06 & 72.26 & 78.12 \\
    & 96  & 32 & 91.55 & 82.71 & 76.16 & 75.13 & 68.39 & 73.51 & 70.68 & 77.13 \\
    \bottomrule
    \end{tabular}
    \label{tab:ratio}
\end{table*}

\subsection{Discussions}

Since we found that a sufficient amount of annotated images are required
 for making the semi-supervised learning effective,
 we further investigated the impact of the mini-batch composition
 on the performance of pose estimation
 by changing the number of annotated images and that of non-annotated images in each mini-batch
 to 32+96, 48+80, 64+64, 80+48, or 96+32.
We used MirrorNet with the HRNet-based pose recognizer $\alpha$~\cite{sun2019deep}
 trained on the LSP dataset, where the ratio of annotated images was set to 20\%.
As shown in \sref{sec:experiment_1},
 the semi-supervised MirrorNet underperformed the supervised MirrorNet
 under the condition of 96+32.

Interestingly, as shown in \tref{tab:ratio},
 the semi-supervised MirrorNet outperformed the supervised MirrorNet
 under the conditions of 32+96, 48+80, 64+64.
In the objective function given by \eref{eq:lowerbound_sum},
 the contributions of annotated and non-annotated images
 are directly affected by the ratio of annotated images in each mini-batch.
Thus, it is necessary to optimize it
 for drawing the full potential of semi-supervised learning.
This should be included in future work.

As the main contribution of our study,
 we proved the concept of the proposed hierarchical mirror system
 in 2D pose estimation for single-person images.
An important future direction of our study
 is to extend MirrorNet to deal with human images
 in which some joints are occluded or out of view.
The noticeable advantage of the fully probabilistic modeling underlying MirrorNet
 is that unobserved joints could be naturally dealt with missing data
 and statistically inferred during the training. 
Besides,
 it is worth extending the current MirrorNet to support 3D pose estimation
 based on the hierarchical mirror system involving the 3D pose VAE at the lower level mirror system.


\section{Conclusion}
\label{sec:conclusion}

Inspired by the cognitive knowledge about the mirror neuron system of humans,
 this paper proposed a deep Bayesian framework called MirrorNet
 for 2D pose estimation from human images.
The key idea is to jointly train the generative models of images and poses
 as well as the recognition models of appearances, scenes, and primitives
 in a fully statistical manner.
From a technical point of view, the two-level mirror systems (VAEs)
 are jointly trained with the hierarchical autoencoding manner
 (image $\rightarrow$ pose $\rightarrow$ primitive $\rightarrow$ pose $\rightarrow$ image),
 such that the plausibility and fidelity of poses are both considered.
Thanks to the nature of the fully generative modeling,
 MirrorNet is the first pose estimation architecture
 that could, in theory, be trained from non-annotated images in an unsupervised manner
 when some appropriate inductive biases are introduced.
We experimentally proved that the whole MirrorNet
 could be jointly trained and outperformed a conventional recognition-model-only method
 in terms of pose estimation performance.
We also showed that the additional use of non-annotated images
 could improve the performance of pose estimation.

The main contribution of this paper
 is that we shed light on the mirror neuron system (or motor theory)
 and build a statistically robust computational model of the human vision system
 by leveraging the expressive power of modern deep Bayesian models.
The same framework can be applied to 3D motion estimation from videos
 by formulating recurrent versions of the pose and image VAEs
 that represent the anatomical plausibility and fidelity of human motions, respectively.
This paper also ushers
 in a new research field of the semi-supervised pose estimation.
We believe that MirrorNet inspires a new approach to multimedia understanding.

\begin{acknowledgements}
This work was supported by the Program for Leading Graduate Schools, ``Graduate Program for Embodiment Informatics'' of the Ministry of Education, Culture, Sports, Science and Technology (MEXT) of Japan, JST ACCEL No.JPMJAC1602, JSPS KAKENHI No.19H04137, and JST-Mirai Program No.JPMJMI19B2.
\end{acknowledgements}



\bibliographystyle{spmpsci}      

\bibliography{main}

\clearpage
\appendix


\section{Lower Bound $\mathcal{L}$}
\label{sec:lowerbound}
\subsection{Variational Lower Bound $\mathcal{L}^\mb{X}$}
Eq. \eqref{eq:derivation_lower_bound_all} is the full derivation of the variational lower bound $\mathcal{L}^\mb{X}$ of $\log p(\mb{X})$ 
 in the unsupervised condition ($\mathcal{L}^\mb{X} = \sum_n \mathcal{L}^\mb{X}_n$).

\subsection{Variational Lower Bound $\mathcal{L}^{\mb{X},\mb{S}}$}
Eq. \eqref{eq:derivation_lower_bound_supervised_all} is the full derivation of the variational lower bound $\mathcal{L}^{\mb{X},\mb{S}}$ of $\log p(\mb{X},\mb{S})$ in the supervised condition ($\mathcal{L}^{\mb{X},\mb{S}} = \sum_n \mathcal{L}^{\mb{X},\mb{S}}_n$).

\begin{widetext}
\begin{align}
 \mathcal{L}^\mb{X}_n
 =&
 \ \mathbb{E}_{q(\mb{s}_n,\mb{a}_n,\mb{g}_n,\mb{z}_n|\mb{x}_n)}
 [\log \left\{ p_\theta(\mb{x}_n|\mb{s}_n,\mb{a}_n,\mb{g}_n) p_\phi(\mb{s}_n|\mb{z}_n) p(\mb{a}_n) p(\mb{g}_n) p(\mb{z}_n) \right\}
 \nonumber\\
 &\:\:\:\:\:\:\:\:\:\:\:\:\:\:\:\:\:\:\:\:\:\:\:\:\:\:\:\:\:\:\:\:\:\:\:\:\:\:
 - \log \left\{q_\alpha(\mb{s}_n|\mb{x}_n) q_\beta(\mb{a}_n|\mb{s}_n,\mb{x}_n) q_\gamma(\mb{g}_n|\mb{s}_n,\mb{x}_n) q_\delta(\mb{z}_n|\mb{s}_n)\right\}]
 \nonumber\\
 =&
 \ \mathbb{E}_{q(\mb{s}_n,\mb{a}_n,\mb{g}_n,\mb{z}_n|\mb{x}_n)}
 [\log p_\theta(\mb{x}_n|\mb{s}_n,\mb{a}_n,\mb{g}_n) + \log p_\phi(\mb{s}_n|\mb{z}_n) + \log p(\mb{a}_n) + \log p(\mb{g}_n) + \log p(\mb{z}_n)
 \nonumber\\
 &\:\:\:\:\:\:\:\:\:\:\:\:\:\:\:\:\:\:\:\:\:\:\:\:\:\:\:\:\:\:\:\:\:\:\:\:\:\:
 - \log q_\alpha(\mb{s}_n|\mb{x}_n) - \log q_\beta(\mb{a}_n|\mb{s}_n,\mb{x}_n) - \log q_\gamma(\mb{g}_n|\mb{s}_n,\mb{x}_n) - \log q_\delta(\mb{z}_n|\mb{s}_n)]
 \nonumber\\
 =&
 \ \mathbb{E}_{q(\mb{s}_n,\mb{a}_n,\mb{g}_n,\mb{z}_n|\mb{x}_n)}
 [\log p_\theta(\mb{x}_n|\mb{s}_n,\mb{a}_n,\mb{g}_n) 
 + \log p_\phi(\mb{s}_n|\mb{z}_n) 
 - \log q_\alpha(\mb{s}_n|\mb{x}_n)
 - (\log q_\beta(\mb{a}_n|\mb{s}_n,\mb{x}_n) - \log p(\mb{a}_n))
 \nonumber\\
 &\:\:\:\:\:\:\:\:\:\:\:\:\:\:\:\:\:\:\:\:\:\:\:\:\:\:\:\:\:\:\:\:\:\:\:\:\:\:
 - (\log q_\gamma(\mb{g}_n|\mb{s}_n,\mb{x}_n) - \log p(\mb{g}_n))
 - (\log q_\delta(\mb{z}_n|\mb{s}_n) - \log p(\mb{z}_n))]
 \nonumber\\
 =&
 \ \mathbb{E}_{q_\alpha(\mb{s}_n|\mb{x}_n) q_\beta(\mb{a}_n|\mb{s}_n,\mb{x}_n) q_\gamma(\mb{g}_n|\mb{s}_n,\mb{x}_n)}
 [\log p_\theta(\mb{x}_n|\mb{s}_n,\mb{a}_n,\mb{g}_n)]
 + \mathbb{E}_{q_\alpha(\mb{s}_n|\mb{x}_n) q_\delta(\mb{z}_n|\mb{s}_n)}
 [\log p_\phi(\mb{s}_n|\mb{z}_n)] 
 \nonumber\\
 &
 - \mathbb{E}_{q_\alpha(\mb{s}_n|\mb{x}_n)}
 [\log q_\alpha(\mb{s}_n|\mb{x}_n)]
 - \mathbb{E}_{q_\alpha(\mb{s}_n|\mb{x}_n)}
 [\mbox{KL}(q_\beta(\mb{a}_n|\mb{s}_n,\mb{x}_n) || p(\mb{a}_n))]
 \nonumber\\
 &
 - \mathbb{E}_{q_\alpha(\mb{s}_n|\mb{x}_n)}
 [\mbox{KL}(q_\gamma(\mb{g}_n|\mb{s}_n,\mb{x}_n) || p(\mb{g}_n))]
 - \mathbb{E}_{q_\alpha(\mb{s}_n|\mb{x}_n)}
 [\mbox{KL}(q_\delta(\mb{z}_n|\mb{s}_n) || p(\mb{z}_n))]
 \nonumber\\
 =&
 - \frac{1}{2} \sum^{D^\mb{x}}_{d_x=1} 
 \mathbb{E}_{q_\alpha(\mb{s}_n|\mb{x}_n) q_\beta(\mb{a}_n|\mb{s}_n,\mb{x}_n) q_\gamma(\mb{g}_n|\mb{s}_n,\mb{x}_n)}
 \left[\log \left(2\pi \sigma^2_{\theta, d_x}(\mb{s}_n, \mb{a}_n, \mb{g}_n) \right) 
 + \frac{\left(\mb{x}_n - \mu_{\theta, d_x}(\mb{s}_n, \mb{a}_n, \mb{g}_n) \right)^2}{\sigma^2_{\theta, d_x}(\mb{s}_n, \mb{a}_n, \mb{g}_n)}\right]
 \nonumber\\
 &
 - \frac{1}{2} \sum^{D^\mb{s}}_{d_s=1}
 \mathbb{E}_{q_\alpha(\mb{s}_n|\mb{x}_n) q_\delta(\mb{z}_n|\mb{s}_n)}
 \left[\log \left(2\pi \sigma^2_{\phi, d_s}(\mb{z}_n) \right) 
 + \frac{\left(\mb{s}_n - \mu_{\phi, d_s}(\mb{z}_n) \right)^2}{\sigma^2_{\phi, d_s}(\mb{z}_n)}\right]
 + \frac{1}{2} \sum^{D^\mb{s}}_{d_s=1}
 \left(1 + \log \left(2\pi \sigma^2_{\alpha, d_s}(\mb{x}_n) \right)\right)
 \nonumber\\
 &
 + \frac{1}{2} \sum^{D^\mb{a}}_{d_a=1} \mathbb{E}_{q_\alpha(\mb{s}_n|\mb{x}_n)} 
 \left[1 + \log (\sigma^2_{\beta, d_a}(\mb{s}_n, \mb{x}_n)) - \mu^2_{\beta, d_a}(\mb{s}_n, \mb{x}_n) - \sigma^2_{\beta, d_a}(\mb{s}_n, \mb{x}_n) \right]
 \nonumber\\
 &
 + \frac{1}{2} \sum^{D^\mb{g}}_{d_g=1} \mathbb{E}_{q_\alpha(\mb{s}_n|\mb{x}_n)} 
 \left[1 + \log (\sigma^2_{\gamma, d_g}(\mb{s}_n, \mb{x}_n)) - \mu^2_{\gamma, d_g}(\mb{s}_n, \mb{x}_n) - \sigma^2_{\gamma, d_g}(\mb{s}_n, \mb{x}_n) \right]
 \nonumber\\
 &
 + \frac{1}{2} \sum^{D^\mb{z}}_{d_z} \mathbb{E}_{q_\alpha(\mb{s}_n|\mb{x}_n)} 
 \left[1 + \log (\sigma^2_{\delta, d_z}(\mb{s}_n)) - \mu^2_{\delta, d_z}(\mb{s}_n) - \sigma^2_{\delta, d_z}(\mb{s}_n) \right].
 \label{eq:derivation_lower_bound_all}
\end{align}

\begin{align}
 \mathcal{L}^{\mb{X},\mb{S}}_n
 =&
 \ \mathbb{E}_{q(\mb{a}_n,\mb{g}_n,\mb{z}_n|\mb{s}_n,\mb{x}_n)}
 [\log \left\{ p_\theta(\mb{x}_n|\mb{s}_n,\mb{a}_n,\mb{g}_n) p_\phi(\mb{s}_n|\mb{z}_n) p(\mb{a}_n) p(\mb{g}_n) p(\mb{z}_n) \right\}
 \nonumber\\
 &\:\:\:\:\:\:\:\:\:\:\:\:\:\:\:\:\:\:\:\:\:\:\:\:\:\:\:\:\:\:\:\:\:\:\:\:\:\:
 - \log \left\{q_\beta(\mb{a}_n|\mb{s}_n,\mb{x}_n) q_\gamma(\mb{g}_n|\mb{s}_n,\mb{x}_n) q_\delta(\mb{z}_n|\mb{s}_n)\right\}]
 \nonumber\\
 =&
 \ \mathbb{E}_{q(\mb{a}_n,\mb{g}_n,\mb{z}_n|\mb{s}_n,\mb{x}_n)}
 [\log p_\theta(\mb{x}_n|\mb{s}_n,\mb{a}_n,\mb{g}_n) + \log p_\phi(\mb{s}_n|\mb{z}_n) + \log p(\mb{a}_n) + \log p(\mb{g}_n) + \log p(\mb{z}_n)
 \nonumber\\
 &\:\:\:\:\:\:\:\:\:\:\:\:\:\:\:\:\:\:\:\:\:\:\:\:\:\:\:\:\:\:\:\:\:\:\:\:\:\:
 - \log q_\beta(\mb{a}_n|\mb{s}_n,\mb{x}_n) - \log q_\gamma(\mb{g}_n|\mb{s}_n,\mb{x}_n) - \log q_\delta(\mb{z}_n|\mb{s}_n)]
 \nonumber\\
 =&
 \ \mathbb{E}_{q(\mb{s}_n,\mb{a}_n,\mb{g}_n,\mb{z}_n|\mb{x}_n)}
 [\log p_\theta(\mb{x}_n|\mb{s}_n,\mb{a}_n,\mb{g}_n) 
 + \log p_\phi(\mb{s}_n|\mb{z}_n) - (\log q_\beta(\mb{a}_n|\mb{s}_n,\mb{x}_n) - \log p(\mb{a}_n)) 
 \nonumber\\
 &\:\:\:\:\:\:\:\:\:\:\:\:\:\:\:\:\:\:\:\:\:\:\:\:\:\:\:\:\:\:\:\:\:\:\:\:\:\:
 - (\log q_\gamma(\mb{g}_n|\mb{s}_n,\mb{x}_n) - \log p(\mb{g}_n))
 - (\log q_\delta(\mb{z}_n|\mb{s}_n) - \log p(\mb{z}_n))]
 \nonumber\\
 =&
 \ \mathbb{E}_{q_\beta(\mb{a}_n|\mb{s}_n,\mb{x}_n) q_\gamma(\mb{g}_n|\mb{s}_n,\mb{x}_n)}
 [\log p_\theta(\mb{x}_n|\mb{s}_n,\mb{a}_n,\mb{g}_n)]
 + \mathbb{E}_{q_\delta(\mb{z}_n|\mb{s}_n)}
 [\log p_\phi(\mb{s}_n|\mb{z}_n)] 
 \nonumber\\
 &
 - \mbox{KL}(q_\beta(\mb{a}_n|\mb{s}_n,\mb{x}_n) || p(\mb{a}_n))
 - \mbox{KL}(q_\gamma(\mb{g}_n|\mb{s}_n,\mb{x}_n) || p(\mb{g}_n))
 - \mbox{KL}(q_\delta(\mb{z}_n|\mb{s}_n) || p(\mb{z}_n))
 \nonumber\\
 =&
 - \frac{1}{2} \sum^{D^\mb{x}}_{d_x=1} 
 \mathbb{E}_{q_\beta(\mb{a}_n|\mb{s}_n,\mb{x}_n) q_\gamma(\mb{g}_n|\mb{s}_n,\mb{x}_n)}
 \left[\log \left(2\pi \sigma^2_{\theta, d_x}(\mb{s}_n, \mb{a}_n, \mb{g}_n) \right) 
 + \frac{\left(\mb{x}_n - \mu_{\theta, d_x}(\mb{s}_n, \mb{a}_n, \mb{g}_n) \right)^2}{\sigma^2_{\theta, d_x}(\mb{s}_n, \mb{a}_n, \mb{g}_n)}\right]
 \nonumber\\
 &
 - \frac{1}{2} \sum^{D^\mb{s}}_{d_s=1}
 \mathbb{E}_{q_\delta(\mb{z}_n|\mb{s}_n)}
 \left[\log \left(2\pi \sigma^2_{\phi, d_s}(\mb{z}_n) \right) 
 + \frac{\left(\mb{s}_n - \mu_{\phi, d_s}(\mb{z}_n) \right)^2}{\sigma^2_{\phi, d_s}(\mb{z}_n)}\right]
 \nonumber\\
 &
 + \frac{1}{2} \sum^{D^\mb{a}}_{d_a=1}
 \left(1 + \log (\sigma^2_{\beta, d_a}(\mb{s}_n, \mb{x}_n)) - \mu^2_{\beta, d_a}(\mb{s}_n, \mb{x}_n) - \sigma^2_{\beta, d_a}(\mb{s}_n, \mb{x}_n) \right)
 \nonumber\\
 &
 + \frac{1}{2} \sum^{D^\mb{g}}_{d_g=1}
 \left(1 + \log (\sigma^2_{\gamma, d_g}(\mb{s}_n, \mb{x}_n)) - \mu^2_{\gamma, d_g}(\mb{s}_n, \mb{x}_n) - \sigma^2_{\gamma, d_g}(\mb{s}_n, \mb{x}_n) \right)
 \nonumber\\
 &
 + \frac{1}{2} \sum^{D^\mb{z}}_{d_z=1}
 \left(1 + \log (\sigma^2_{\delta, d_z}(\mb{s}_n)) - \mu^2_{\delta, d_z}(\mb{s}_n) - \sigma^2_{\delta, d_z}(\mb{s}_n) \right).
 \label{eq:derivation_lower_bound_supervised_all}
\end{align}
\end{widetext}

\begin{figure*}[ht]
    \centering
    \begin{tabular}{@{}c|ccccc@{}}
        \toprule
        \hfil Input \hfil & \multicolumn{5}{c}{Reconstruction}  \\
        \midrule
        & \multicolumn{5}{c}{The number of annotation images used for training}  \\
        & 2200 & 4400 & 6600 & 8800 & 11000 \\
        \midrule

            \begin{minipage}{7.6em}
                \centering

                {\includegraphics[width=\linewidth]{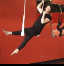}}
            \end{minipage} &
            \begin{minipage}{7.6em}
                \centering

                {\includegraphics[width=\linewidth]{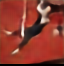}}
            \end{minipage} &
            \begin{minipage}{7.6em}
                \centering

                {\includegraphics[width=\linewidth]{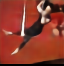}}
            \end{minipage} &
            \begin{minipage}{7.6em}
                \centering

                {\includegraphics[width=\linewidth]{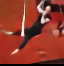}}
            \end{minipage} &
            \begin{minipage}{7.6em}
                \centering

                {\includegraphics[width=\linewidth]{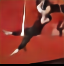}}
            \end{minipage} &
            \begin{minipage}{7.6em}
                \centering

                {\includegraphics[width=\linewidth]{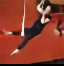}}
            \end{minipage} \\
        \midrule

            \begin{minipage}{7.6em}
                \centering

                {\includegraphics[width=\linewidth]{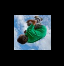}}
            \end{minipage} &
            \begin{minipage}{7.6em}
                \centering

                {\includegraphics[width=\linewidth]{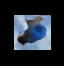}}
            \end{minipage} &
            \begin{minipage}{7.6em}
                \centering

                {\includegraphics[width=\linewidth]{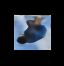}}
            \end{minipage} &
            \begin{minipage}{7.6em}
                \centering

                {\includegraphics[width=\linewidth]{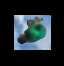}}
            \end{minipage} &
            \begin{minipage}{7.6em}
                \centering

                {\includegraphics[width=\linewidth]{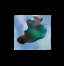}}
            \end{minipage} &
            \begin{minipage}{7.6em}
                \centering

                {\includegraphics[width=\linewidth]{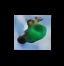}}
            \end{minipage} \\
        \midrule

            \begin{minipage}{7.6em}
                \centering

                {\includegraphics[width=\linewidth]{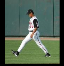}}
            \end{minipage} &
            \begin{minipage}{7.6em}
                \centering

                {\includegraphics[width=\linewidth]{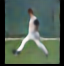}}
            \end{minipage} &
            \begin{minipage}{7.6em}
                \centering

                {\includegraphics[width=\linewidth]{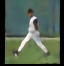}}
            \end{minipage} &
            \begin{minipage}{7.6em}
                \centering

                {\includegraphics[width=\linewidth]{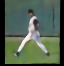}}
            \end{minipage} &
            \begin{minipage}{7.6em}
                \centering

                {\includegraphics[width=\linewidth]{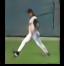}}
            \end{minipage} &
            \begin{minipage}{7.6em}
                \centering

                {\includegraphics[width=\linewidth]{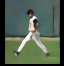}}
            \end{minipage} \\
        \midrule

            \begin{minipage}{7.6em}
                \centering

                {\includegraphics[width=\linewidth]{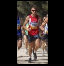}}
            \end{minipage} &
            \begin{minipage}{7.6em}
                \centering

                {\includegraphics[width=\linewidth]{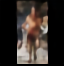}}
            \end{minipage} &
            \begin{minipage}{7.6em}
                \centering

                {\includegraphics[width=\linewidth]{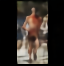}}
            \end{minipage} &
            \begin{minipage}{7.6em}
                \centering

                {\includegraphics[width=\linewidth]{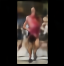}}
            \end{minipage} &
            \begin{minipage}{7.6em}
                \centering

                {\includegraphics[width=\linewidth]{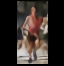}}
            \end{minipage} &
            \begin{minipage}{7.6em}
                \centering

                {\includegraphics[width=\linewidth]{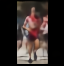}}
            \end{minipage} \\
        \bottomrule
        \end{tabular}

    \caption{
      Reconstruction of human images based on the pose-conditioned image VAE.
      The LSP dataset \cite{johnson2010clustered,johnson2011learning} was used for training.
      A larger amount of annotation images resulted in a better quality of generated images
    }
    \label{fig:genimage}
\end{figure*}

\newlength{\appendixeachimagewidth}
\setlength{\appendixeachimagewidth}{6em}

\newcommand{\figgenpose}{%
    \begin{tabular}{@{}ccc@{}}
        \toprule
        Original & Input & Reconst. \\
        \midrule
            \begin{minipage}{\appendixeachimagewidth}
                \centering

                {\includegraphics[width=\linewidth]{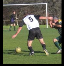}}
            \end{minipage} &
            \begin{minipage}{\appendixeachimagewidth}
                \centering

                {\includegraphics[width=\linewidth]{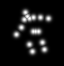}}
            \end{minipage} &
            \begin{minipage}{\appendixeachimagewidth}
                \centering

                {\includegraphics[width=\linewidth]{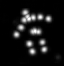}}
            \end{minipage} 
 \\
        \midrule
            \begin{minipage}{\appendixeachimagewidth}
                \centering

                {\includegraphics[width=\linewidth]{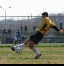}}
            \end{minipage} &
            \begin{minipage}{\appendixeachimagewidth}
                \centering

                {\includegraphics[width=\linewidth]{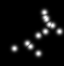}}
            \end{minipage} &
            \begin{minipage}{\appendixeachimagewidth}
                \centering

                {\includegraphics[width=\linewidth]{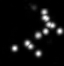}}
            \end{minipage} 
 \\
        \midrule
            \begin{minipage}{\appendixeachimagewidth}
                \centering

                {\includegraphics[width=\linewidth]{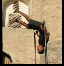}}
            \end{minipage} &
            \begin{minipage}{\appendixeachimagewidth}
                \centering

                {\includegraphics[width=\linewidth]{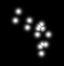}}
            \end{minipage} &
            \begin{minipage}{\appendixeachimagewidth}
                \centering

                {\includegraphics[width=\linewidth]{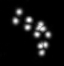}}
            \end{minipage} 
 \\
        \bottomrule
    \end{tabular}
}

\newcommand{\figgenmask}{%
    \begin{tabular}{@{}cc|cc@{}}
        \toprule
        \multicolumn{2}{c|}{Input} & Output & Ground truth \\
        \midrule
            \begin{minipage}{\appendixeachimagewidth}
                \centering

                {\includegraphics[width=\linewidth]{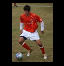}}
            \end{minipage} &
            \begin{minipage}{\appendixeachimagewidth}
                \centering

                {\includegraphics[width=\linewidth]{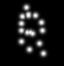}}
            \end{minipage} &
            \begin{minipage}{\appendixeachimagewidth}
                \centering

                {\includegraphics[width=\linewidth]{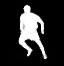}}
            \end{minipage} &
            \begin{minipage}{\appendixeachimagewidth}
                \centering

                {\includegraphics[width=\linewidth]{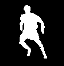}}
            \end{minipage} \\
        \midrule
            \begin{minipage}{\appendixeachimagewidth}
                \centering

                {\includegraphics[width=\linewidth]{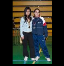}}
            \end{minipage} &
            \begin{minipage}{\appendixeachimagewidth}
                \centering

                {\includegraphics[width=\linewidth]{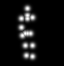}}
            \end{minipage} &
            \begin{minipage}{\appendixeachimagewidth}
                \centering

                {\includegraphics[width=\linewidth]{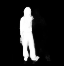}}
            \end{minipage} &
            \begin{minipage}{\appendixeachimagewidth}
                \centering

                {\includegraphics[width=\linewidth]{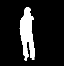}}
            \end{minipage} \\
        \midrule
            \begin{minipage}{\appendixeachimagewidth}
                \centering

                {\includegraphics[width=\linewidth]{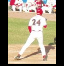}}
            \end{minipage} &
            \begin{minipage}{\appendixeachimagewidth}
                \centering

                {\includegraphics[width=\linewidth]{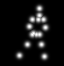}}
            \end{minipage} &
            \begin{minipage}{\appendixeachimagewidth}
                \centering

                {\includegraphics[width=\linewidth]{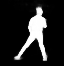}}
            \end{minipage} &
            \begin{minipage}{\appendixeachimagewidth}
                \centering

                {\includegraphics[width=\linewidth]{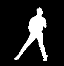}}
            \end{minipage} \\
        \bottomrule
    \end{tabular}
}

\begin{figure*}[ht]
    \begin{tabular*}{\linewidth}{@{}p{.42\linewidth}@{\extracolsep{\fill}}p{.54\linewidth}@{}}
        \begin{minipage}{\linewidth}
            \centering
            \figgenpose
            \caption{Reconstruction of 16-joint heatmaps based on the pose VAE.
            For the purpose of visualization, 16 heatmaps are superimposed to a single heatmap}
            \label{fig:genpose}
        \end{minipage}
        &
        \begin{minipage}{\linewidth}
            \centering
            \figgenmask
            \caption{Prediction of silhouette images (foreground mask images) 
            from 16 joint heatmaps based on the mask generator $\phi$.
            The ground truth images are taken from \cite{Lassner:up2017}}
            \label{fig:genmask}
        \end{minipage}
    \end{tabular*}
\end{figure*}

\end{document}